\documentclass{article}
\usepackage[utf8]{inputenc}
\usepackage{graphicx}
\usepackage{amsthm}
\usepackage{amsmath, amssymb, amsthm}
\usepackage{amsfonts}
\usepackage{bm}
\usepackage[margin=1in]{geometry}
\usepackage{enumerate}
\usepackage[shortlabels]{enumitem}
\usepackage{mathtools}
\usepackage{blkarray}
\usepackage[english]{babel}      
\usepackage{indentfirst}
\usepackage{subcaption}
\usepackage{csquotes}
\usepackage{tikz}
\usetikzlibrary{patterns}
\usepackage{tikz-cd}
\usepackage{url}
\usepackage{color}
\usepackage{setspace,lipsum}

\newcommand{\beginsupplement}{%
        \setcounter{table}{0}
        \renewcommand{\thetable}{S\arabic{table}}%
        \setcounter{figure}{0}
        \renewcommand{\thefigure}{S\arabic{figure}}%
     }

\definecolor{sion}{RGB}{147,99,145}

\title{Architecture and evolution of semantic networks in mathematics texts}

\author{Nicolas H. Christianson$^{1, 2}$, Ann Sizemore Blevins$^2$, Danielle S. Bassett$^{2-7,*}$}

\date{%
    $^1$John A. Paulson School of Engineering and Applied Sciences, Harvard University, Cambridge, MA 02138, USA\\%
    $^2$Department of Bioengineering, School of Engineering \& Applied Science, University of Pennsylvania, Philadelphia, PA 19104, USA\\%
    $^3$Department of Physics \& Astronomy, College of Arts \& Sciences, University of Pennsylvania, Philadelphia, PA 19104, USA\\%
    $^4$Department of Electrical \& Systems Engineering, School of Engineering \& Applied Science, University of Pennsylvania, Philadelphia, PA 19104, USA\\%
$^5$Department of Neurology, Perelman School of Medicine, University of Pennsylvania, Philadelphia, PA 19104, USA\\%
$^6$Department of Psychiatry, Perelman School of Medicine, University of Pennsylvania, Philadelphia, PA 19104, USA\\%
$^7$Santa Fe Institute, Santa Fe, NM 87501, USA\\%
$^*$dsb@seas.upenn.edu\\[2ex]%
    \today
}

\begin{document}
\maketitle

\singlespacing
\noindent \textbf{Abstract.} Knowledge is a network of interconnected concepts. Yet, precisely how the topological structure of knowledge constrains its acquisition remains unknown, hampering the development of learning enhancement strategies. Here we study the topological structure of semantic networks reflecting mathematical concepts and their relations in college-level linear algebra texts. We hypothesize that these networks will exhibit structural order, reflecting the logical sequence of topics that ensures accessibility. We find that the networks exhibit strong core-periphery architecture, where a dense core of concepts presented early is complemented with a sparse periphery presented evenly throughout the exposition; the latter is composed of many small modules each reflecting more narrow domains. Using tools from applied topology, we find that the expositional evolution of the semantic networks produces and subsequently fills knowledge gaps, and that the density of these gaps tracks negatively with community ratings of each textbook. Broadly, our study lays the groundwork for future efforts developing optimal design principles for textbook exposition and teaching in a classroom setting.

\clearpage
\newpage 

\section*{Introduction}

Knowledge has been distilled into formal representations for millennia \cite{jsowa, sep-epistemology}. Such efforts have sought to explain human reasoning and support artificial intelligence \cite{hartley, lehmann, nickel}. Semantic networks organize information by detailing concepts and their relations as the nodes and edges of a graph \cite{steyvers}. In an educational context, concept maps reflect students' understanding of information in a similar manner, but may be used to evaluate comprehension \cite{gallenstein,broggy2009integrating,hill_concept,daley_concept} and identify topics that are most difficult to connect to other concepts \cite{lapp}. With the capacity to construct semantic networks, concept maps, and similar formal representations of knowledge comes the challenge of distilling mechanisms of knowledge acquisition.

Network science offers an appropriate conceptual language and useful mathematical toolset with which to meet this challenge \cite{newman2010networks}. In the parlance of network science, semantic networks of language tend to exhibit highly ordered architectures with strong local clustering, relatively short paths between any pair of nodes, and a few hubs, which are connected to an unexpectedly large number of other nodes \cite{steyvers, utsumi}. Recent work using highly stylized laboratory experiments provides some preliminary evidence that network structure may play a role in how humans process information \cite{lynn2019human} and acquire knowledge \cite{costa, karuza, koponen}. Yet extending these findings to the real world has proven difficult and it remains unknown precisely how the network structure of knowledge in the form of science textbooks \cite{textbook_networks}, science and mathematics topics on Wikipedia \cite{Fang2011WikipediaAD}, and even formal scientific papers \cite{lucychai,PEREIRA20111192} impacts the learnability of these content domains. Furthermore, as learning is a process, studies of semantic network architecture would benefit from evaluating a network's dynamic structure as it unfurls over the course of presentation, exposition, or acquisition. The education literature establishes that the order in which topics are introduced can help or hinder learning at this level \cite{ritter,kapur}, but a rigorous understanding of order and dynamic structure in knowledge acquisition has not been formalized in ecologically valid experimental settings. 

Here we seek to address these limitations by studying semantic networks of mathematical concepts in linear algebra textbooks \cite{mowat2008,Mowat2010}. A common college-level course, the subject is rigorous and logical, sequentially introducing concepts that naturally relate to, depend on, and follow from other concepts. To begin, we seek to understand the structure of these inter-concept relations in textbooks, which present the knowledge in a thoughtfully ordered and comprehensive exposition. While each author may introduce and relate topics in a different order, we assume that each text serves to elucidate and approximate the latent structure of the domain of knowledge it conveys. Using techniques from network science \cite{newman2010networks}, we test the hypothesis that these semantic networks exhibit structural order, indicating a logical sequence of topics that ensures accessibility. Motivated by a recent report that language acquisition proceeds through an ordered progression filling knowledge gaps \cite{sizemore}, we use \textit{persistent homology} \cite{carlsson,zomorodian, Edelsbrunner2013PersistentHT, Otter2017} to track the growth and development of topological cavities in the semantic network. We predict that fewer knowledge gaps will exist in the texts than in null models of randomly growing semantic networks; withholding connections between topics that have already been taught is unlikely to effectively convey knowledge. Finally, we compare the growth of semantic networks elicited from multiple texts, in terms of their different expositional structures and topic orderings. We hypothesize that the degree to which knowledge gaps are created and persist within texts may be related to the complexity or difficulty of a text, and to the knowledge it conveys. Broadly, our quantitative evaluation of the differing structures and expositional layouts of distinct textbooks provides a foundation for future work examining the effects of topic ordering and network architecture on classroom learning.

\section*{Results}
\begin{figure}
	\centering
	\includegraphics[width=0.6\linewidth]{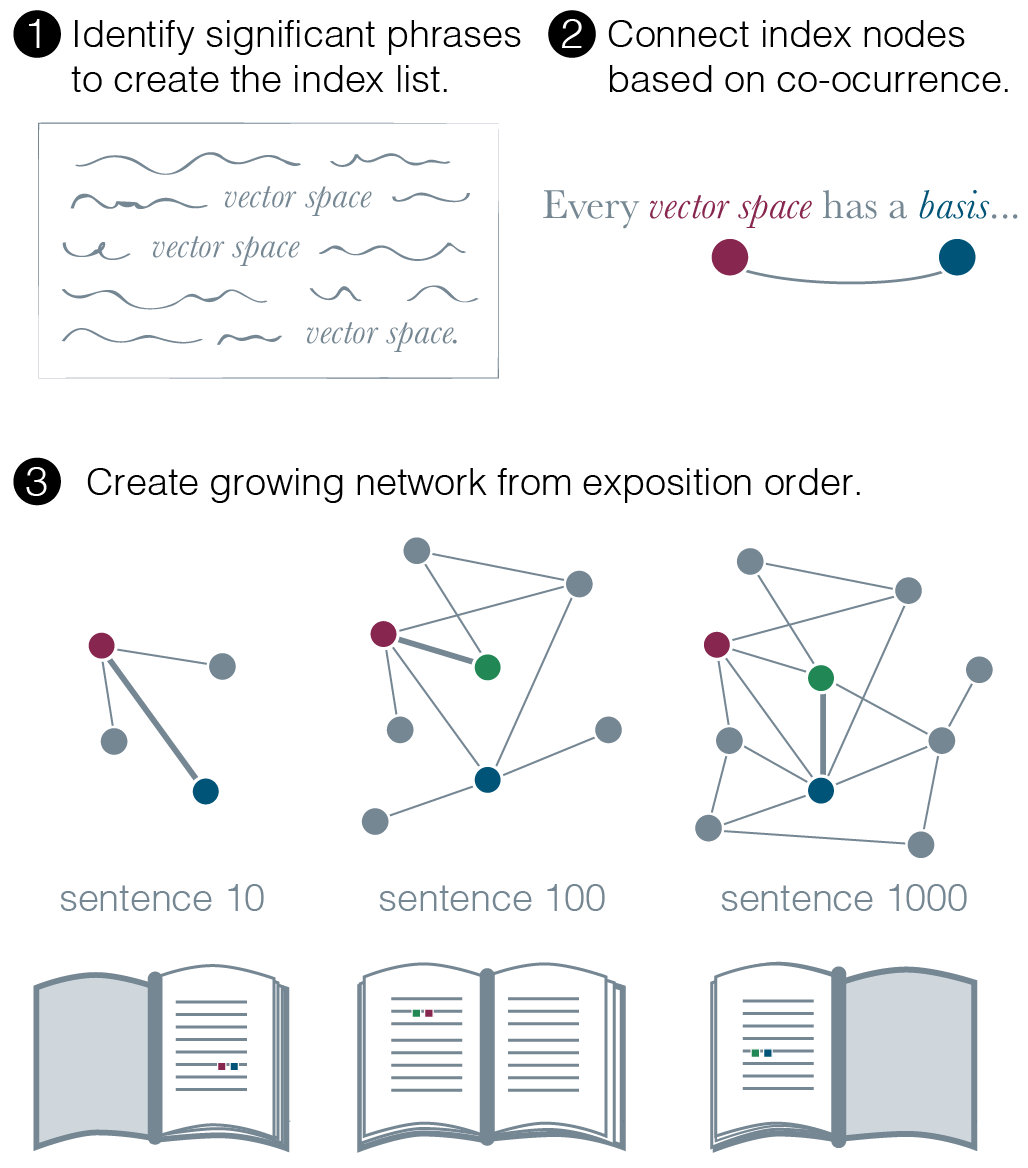}
	\caption{Extracting growing semantic networks from textbooks. \emph{(1)} The index set is populated with phrases conveying significant mathematical concepts. \emph{(2)} Any index nodes that co-occur within the same sentence are connected by an edge. \emph{(3)} This procedure is applied to each sentence in the exposition, forming a view of the semantic network as it grows throughout the text.}
	\label{fig0}
\end{figure}

We constructed semantic networks and expositional growing networks from 10 linear algebra textbooks (see Methods). We first used a modified version of the RAKE algorithm \cite{RAKE} to identify significant phrases (Fig.~\ref{fig0}, step 1), which we refer to collectively as the index list of concepts. We represent these concepts as nodes, and connect two nodes by an edge if their corresponding concepts co-occur within the same sentence (Fig.~\ref{fig0}, step 2). To mimic the growth of a reader's knowledge network, we add nodes and edges as soon as they are mentioned in the book (Fig.~\ref{fig0}, step 3). Across textbooks, node sets ranged in length from 146 to 453 (average 279.4) and edge densities ranged from 0.0748 to 0.204 (average 0.129). In what follows, we characterize the semantic network growth of all texts, and when useful we give examples from individual texts referred to by author last name.

\subsection*{Meso-scale structure of semantic networks}

\begin{figure}[t]
	\centering
	\includegraphics[width=0.6\linewidth]{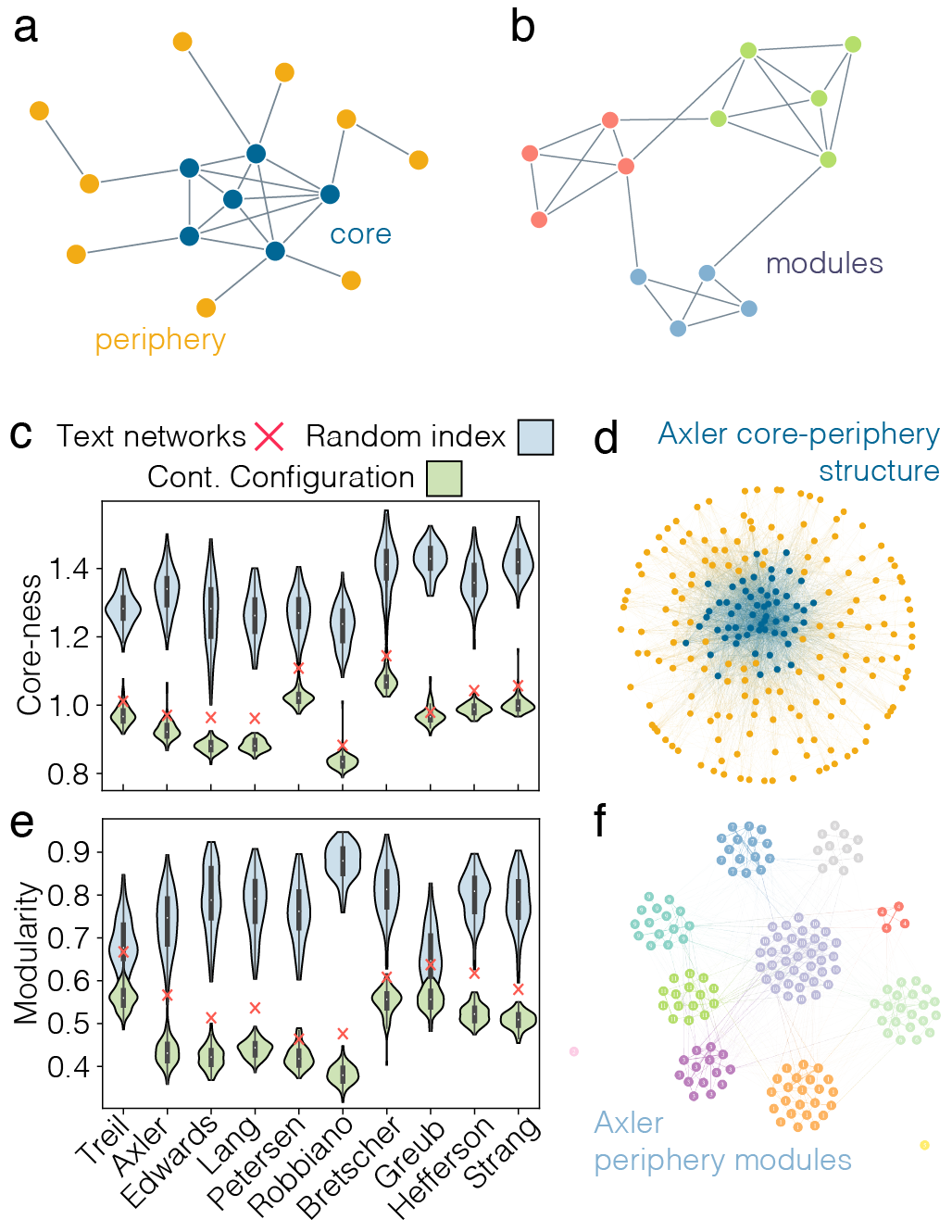}
	\caption{Meso-scale structure of semantic networks. \emph{(a)} A schematic of core-periphery structure, with densely-connected core nodes and a sparsely-connected periphery. \emph{(b)} A schematic of community structure, with densely-connected communities which are themselves sparsely connected to each other. \emph{(c)} The core-ness statistic of each network and of the corresponding random index and continuous configuration null ensembles. \emph{(d)} Visualization of the Axler core-periphery structure. \emph{(e)} Modularity statistics of the periphery of each network and of the corresponding random index and continuous configuration null ensembles. \emph{(f)} Visualization of the Axler periphery community structure. Graph visualizations generated with Graph-tool \cite{peixoto_graph-tool_2014}. See Supplementary Table~\ref{table:comm_concepts} for example nodes present in the Axler periphery communities.}
	\label{fig1}
\end{figure}

Mathematics as a field and linear algebra as a subject contain many fundamental topics and conceptual connections between those topics. Practitioners and authors might contest which topics are fundamental, and which are more tangential, or less strongly linked to the rest. Within a network, this organizational scheme can manifest as core-periphery structure where fundamental concepts are densely connected to one another, while peripheral concepts connect to the core but not to one another (Fig.~\ref{fig1}a). To assess this structure in a semantic network constructed from the whole text, we calculate the core-periphery statistic and compare statistic values to those obtained from two null models (Fig.~\ref{fig1}c): (i) a \emph{random index null model}, in which random words from each text are used to generate an expositional network, and (ii) a \emph{continuous configuration model}, in which the original network is rewired while maintaining node degree and strength. Generally, we observe that the empirical semantic networks show greater core-periphery organization than the continuous configuration model, suggesting the presence of a strongly connected core of topics along with a set of sparsely connected periphery topics given the degree and strength distributions. Interestingly, we also observe that the empirical networks show less core-periphery organization than the random index model, indicating that the networks of math terms are more homogeneous than a network of randomly chosen words.

We next investigate the internal structure of the core and periphery. For the core, we find that across texts many similar words participate, including `determinant', `vector space', and `matrix' as expected (see Supplementary Table~\ref{table:core_freqs}). In contrast, we expect that the periphery contains terms more specific to a given book and its particular sub-topics. We therefore hypothesize that the periphery will display community structure (Fig.~\ref{fig1}d). To test our hypothesis, we calculate the modularity of the periphery subnetwork, along with the relevant subnetworks of the random index and continuous configuration null models. We observe that the periphery of each semantic network generally exhibits a modular organization that is stronger than that of the continuous configuration model, but weaker than that of the random index model (Fig.~\ref{fig1}e). Intuitively, while randomly chosen words may display strong modularity due to greater variation in semantic relationships and frequencies, mathematics phrases are used in a more modular fashion than expected from the rewired continuous configuration model, perhaps due to the nature of focusing on one general idea at a time in chapters and sections.

\subsection*{Expositional development of the large-scale structure}

\begin{figure}
	\centering
	\includegraphics[width=0.6\linewidth]{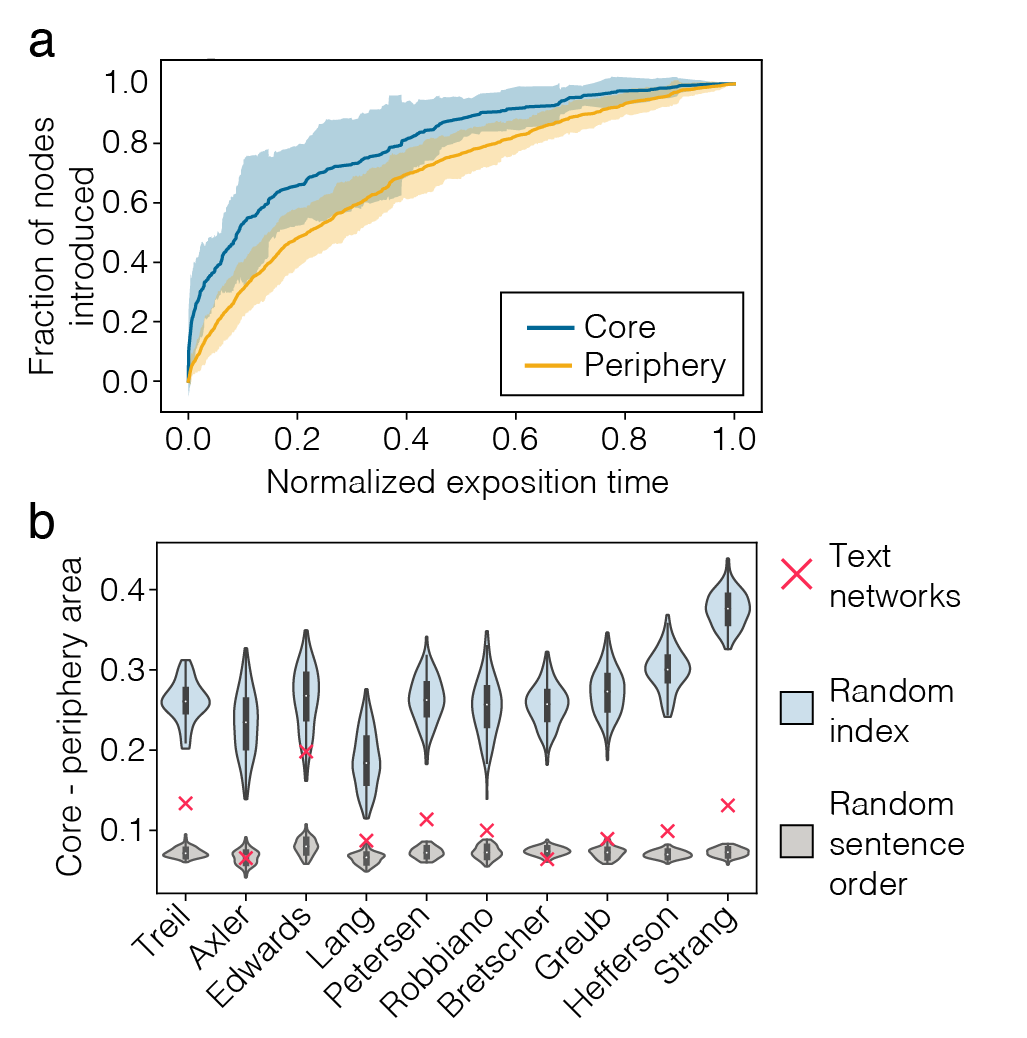}
    \caption{Development of core-periphery structure during exposition. \emph{(a)} Core and periphery development curves, showing the fraction of nodes in each group introduced by a particular time in the exposition; mean $\pm$ 2 standard deviations across all texts. \emph{(b)} Difference in area between core and periphery development curves for all texts and corresponding random index and random sentence order null ensembles.}
    \label{fig2}
\end{figure}

How does the identified network structure develop along a text's exposition? We find that the expositional introduction of nodes in the final network's core precedes the introduction of periphery nodes throughout the exposition (Fig.~\ref{fig2}a). We quantify this observation by calculating the area between the core and periphery node introduction curves; high values indicate that the core appears much earlier than the periphery, and low values indicate that the core and periphery appear at a more equal rate. The areas range from 0.064 and 0.20 across texts, and a one-sample $t$-test rejects the null hypothesis that these values are drawn from a distribution with mean 0 ($t=8.65$, $p=1.18 \times 10^{-5}$; calculated with the SciPy library, version 1.1.0 \cite{scipy}). 

Next, we compare the areas obtained from the texts to those expected in statistical null models. Notably, we find that the empirical periphery is introduced earlier (relative to the core) than expected from the random index model, which has a more stark difference between core and periphery introduction (Fig.~\ref{fig2}b). We observe no consistent trend across texts in comparison to a \emph{random sentence model}, in which we use the original index list to build a growing graph from the texts after randomizing sentence order. While many texts show a marked discrepancy between the core and periphery development, others show a more even development. These differences across texts could reflect different expositional styles amongst different authors: some may choose to introduce core topics initially and save extra tangents for later, while others may involve discussions of peripheral topics throughout the text for motivation. Additionally, we take a similar approach in examining the relative rate of introduction of edges connecting different types of groups within the core and periphery, and find that of all edge types, those connecting concepts within a single periphery community are introduced the most sporadically, with some communities being fully introduced early on in the text, and some being introduced later (see Supplementary Fig.~\ref{fig:supp_edgeintro}).

\subsection*{Expositional development of knowledge gaps}

\begin{figure}[t]
	\centering
	\includegraphics[width=0.6\linewidth]{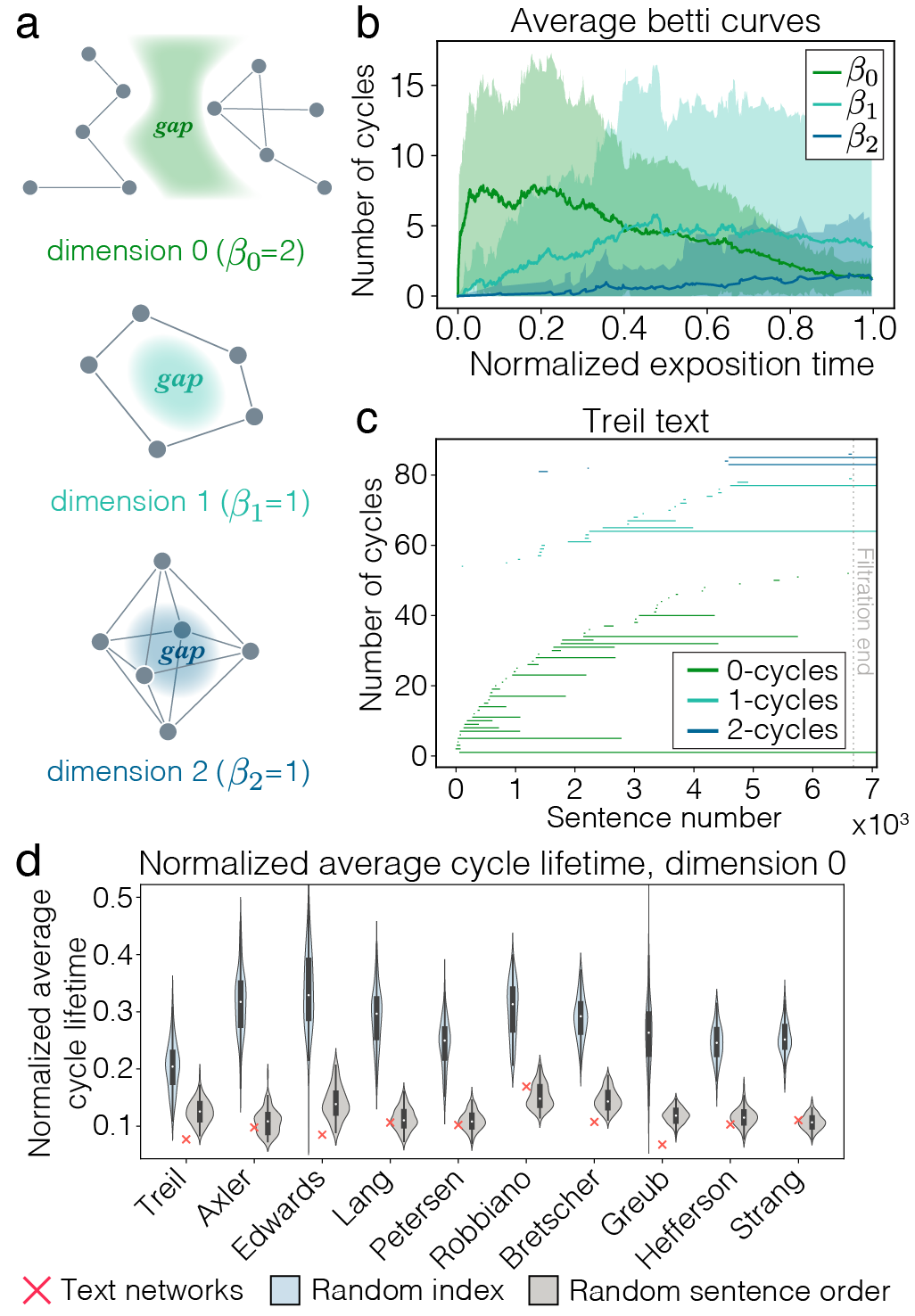}
    \caption{Development and persistence of knowledge gaps throughout exposition. \emph{(a)} Examples of knowledge gaps in dimensions 0, 1, and 2. \emph{(b)} Number of live knowledge gaps, or cycles, in each dimension throughout exposition; mean $\pm$ 2 standard deviations across all texts. \emph{(c)} Barcode for the Treil text, showing introduction, persistence, and death of cycles introduced throughout exposition. Barcodes for other texts are provided in Supplementary Figs.~\ref{fig:sbb_1}, \ref{fig:sbb_2}, \ref{fig:bb_1}, \ref{fig:bb_2}. \emph{(d)} The 0-dimensional normalized average cycle lifetime across all texts, as well as corresponding random index and random sentence order null models.}
    \label{fig3}
\end{figure}

In studying core and periphery formation, we focused on densely connected areas in the growing networks; now we turn to a study of sparsely connected areas. Specifically, we seek to understand how voids or knowledge gaps might emerge and evolve throughout the exposition. Teaching strategies may intentionally leave open a connection or an area of the knowledge space in order to more intuitively reveal the connection later when a learner has more experience, or to provide the reader the opportunity to derive the connection on his/her/their own. A lack of connections between concepts can manifest as a topological gap in the network (Fig.~\ref{fig3}a). 

To detect gaps that form and evolve throughout the text, we compute the persistent homology \cite{carlsson, zomorodian, Edelsbrunner2013PersistentHT, Otter2017} of the ordered set of networks composed of nodes and edges that exist at each point in the exposition; note, this ordered set of binary graphs is referred to as a filtration. We specifically detect gaps between connected components (dimension 0 homology $>1$), cavities within rings of edges (dimension 1 homology $>0$), or voids within polyhedra (dimension 2 homology $> 0$). We say that these so-called persistent cavities are \emph{born} at the first instance of their appearance in the network, they \emph{live} as long as the network grows and the topological void still persists, and they \emph{die} when they are either connected to another previously disconnected component (in the case of dimension 0) or are tessellated by crossing edges (in the higher dimension cases). We invite the reader to refer to the Methods for a more rigorous description of persistent homology in this application.

In order to detect emerging and evolving gaps throughout exposition, we compute the persistent homology of each text. The number of gaps of dimension $n$ that are alive at a given point in the filtration, called the Betti curve, is denoted $\beta_n$. We see that the texts tend to generate a large number of components, as manifest by the initial $\beta_0$ peak, followed by a rise in $\beta_1$, and finally a slow and steady increase in $\beta_2$ (Fig.~\ref{fig3}b). For each text, we summarize the life and death of each persistent gap in a barcode (Fig.~\ref{fig3}c). Each bar represents a single persistent cavity; the left endpoint of the bar indicates the birth time of the persistent cavity, while the right endpoint indicates the death time. Across all texts, we see that although many persistent cavities are killed soon after birth, a non-trivial number of gaps in each of the three dimensions persist throughout many sentences, suggesting that long-lived gaps are a consistent hallmark of the growing text structure. 

To further evaluate the substantiveness of the gap architecture, we compared the persistent homology of the text networks to two filtration-based null models. In the first null model, we use the introduction of concepts to order the complete network for the filtration. More precisely, this node-ordered null model adds a node at the first mention of the concept, and also adds all of the connections that will ever exist from that node to previously acquired concepts. This model mimics the teaching strategy of introducing all connections of a new concept to anything previously taught. We find that the node-ordered model produces almost no persistent homology, in stark contrast to the original text (see Supplementary Figs.~\ref{fig:bb_1}, \ref{fig:bb_2}). This result suggests that the text expositions consistently leave connections between already-learned concepts for later discussion. We use a second null model to determine whether the totally random introduction of edges might produce similar progressions of persistent cavities. We find that this random edge order model produces an order of magnitude more persistent cavities of dimension 1 and 2 than the original text (see Supplementary Figs.~\ref{fig:bb_1}, \ref{fig:bb_2}). Broadly, the presence of a few long-lived cavities in the actual text are consistent with the notion that knowledge gaps exist but are introduced sparingly, and that introducing connections to all topics previously learned is not the strategy of these texts.

At this point we know that throughout the text the introduction of terms and connections forms and fills gaps as a reader progresses. However, we do not yet know if the number and longevity of persistent cavities is different than we would expect from any growing semantic network in the text or from a reordered text. In order to answer this question, we define the normalized average cycle lifetime in dimension $n$ as the sum of all persistent cavity lifetimes normalized by the number of cavities and filtration length (similar to metrics defined in \cite{adcock}, see Methods for details). Then intuitively a large value of normalized average cycle lifetime suggests that multiple long-lived persistent gaps exist, while a small value suggests that any gaps that form will die shortly after birth. We show the distributions of normalized average cycle lifetime values in dimension 0 in Fig.~\ref{fig3}d for the random index and random sentence models, and the corresponding distributions for dimensions 1 and 2 in Supplementary Figs.~\ref{fig:sentbarcodedens}, \ref{fig:oaatbarcodedens}. For completeness, we also include the barcodes and Betti curves for each model in the Supplementary Figs.~\ref{fig:sbb_1}, \ref{fig:sbb_2}, \ref{fig:bb_1}, \ref{fig:bb_2}. Strikingly, the original text expositions generally fall below both null models' expected normalized average cycle lifetimes in dimension 0. This observation suggests that the exposition proceeds in a manner that may intentionally avoid developing disconnected topics, or possibly connects new topics to others very quickly. In dimensions 1 and 2, texts' normalized average cycle lifetimes vary more in relation to their null models, with only a handful of texts showing lower values than the null models. 

\subsection*{Evolving structure and text properties}

After characterizing the structural features of the growing text networks, we next ask if these features might relate to text rating. Perhaps some readers particularly enjoy a book that leaves open many gaps motivating future study, while others enjoy a book with a stronger core offering conceptual closure. To determine whether readers' preferences relate to network structure, we used average text rating across all editions from Goodreads (\textit{goodreads.com}). We kept any text which had at least five ratings, which was the case for seven of the ten texts. We observe no significant correlation between average text rating and normalized average cycle lifetime across texts' sentence-based filtrations (Fig.~\ref{fig4}a; see Supplementary Table~\ref{table:corr_pvalues} for all Spearman's correlation coefficients and $p$-values). We also consider a \textit{one-at-a-time} (OAAT) filtration (see Supplementary Methods), which in addition to allowing for comparability in persistent homology across texts and null models, provides additional information not just about a text's knowledge gaps on the sentence scale, but furthermore its sub-sentence topological structure. Remarkably, we observe significant negative correlations between average rating and OAAT normalized average cycle lifetime in dimensions 0 (Spearman's correlation coefficient $\rho=-0.857$, $p=0.0137$) and 2 ($\rho=-0.893$, $p=0.00681$), as well as the mean cycle lifetime averaged over dimensions 0, 1, and 2 ($\rho=-0.821$, $p=0.0234$) (Fig.~\ref{fig4}b, see Supplementary Table~\ref{table:corr_pvalues} for all statistics). Intuitively, these results provide preliminary support for the notion that the extent of knowledge gaps in exposition influences the quality of a text as a learning tool. However, these results only account for seven of the ten books, due to lack of availability of ratings for the others; as such, further work will be necessary to confirm the reliability of these findings in larger samples. For a description of additional relationships between the texts' structural features and broader text characteristics, we refer the reader to the Supplement.

\begin{figure}[t]
	\centering
	\includegraphics[width=0.6\linewidth]{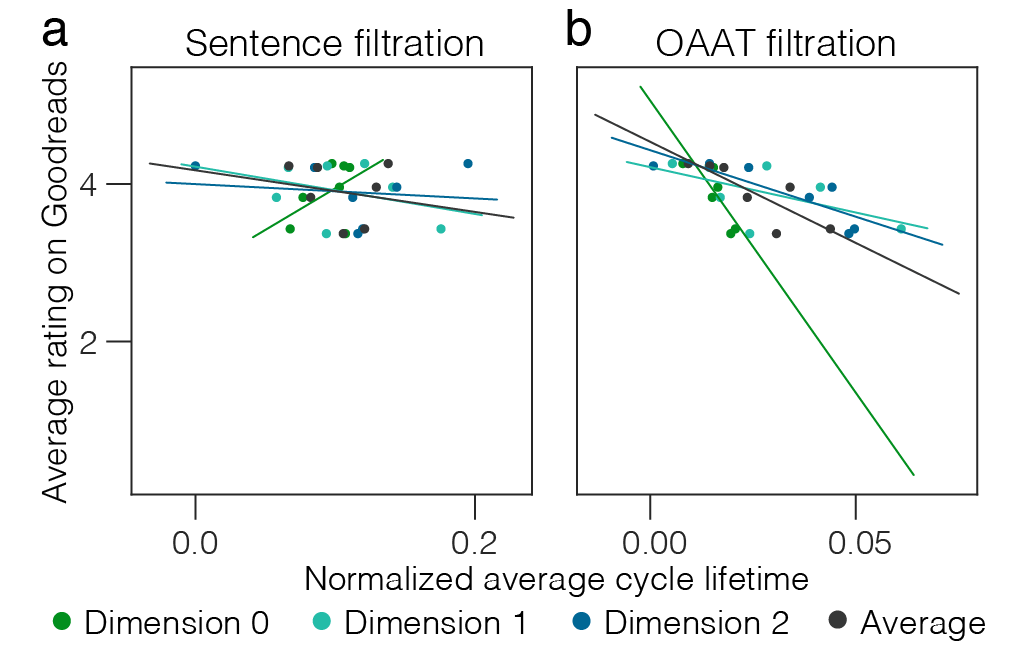}
	\caption{Relation between textbook ratings and network topology. Scatterplots and best-fit lines for average Goodreads rating versus the normalized average cycle lifetime in dimensions 0 through 2 and the average over dimensions, across all texts for the \emph{(a)} sentence filtrations and \emph{(b)} OAAT filtrations.}
	\label{fig4}
\end{figure}

\section*{Discussion}

Here we examined the structure and topological development of semantic networks of mathematical knowledge as extracted from linear algebra texts. Meso-scale structural analysis indicates that the semantic networks exhibit strong core-periphery structure, where a tightly knit group of concepts form a core, surrounded by sparsely-connected periphery concepts that are grouped into communities. Furthermore, these features appear to relate to the growth of the networks over the course of exposition; the cores of networks are built more quickly than the peripheries, and edges within each particular periphery community are introduced at varied times over the course of exposition. Using persistent homology, we extracted the knowledge gaps inherent in the exposition and found that the number of distinct connected components tends to decrease throughout the text, while topological cavities tend to increase. Finally, we examined possible relationships between the extent and persistence of knowledge gaps and other features of a text and its associated semantic network, providing motivation for future work examining the role of knowledge gaps in learning.

\subsection*{Structure and evolution of mesoscale features in semantic networks}

The prevalence of core-periphery and community structures in the networks we examine is consistent with a hierarchical structuring of mathematical knowledge, in which there exist a set of foundational concepts (the core), which are necessary for the subsequent logical development of subsidiary (periphery) concepts, which themselves are hierarchically organized into related communities. The generic notion of hierarchical structure in mathematics has been discussed in the context of presenting a logical sequence of concepts in education \cite{hierarchy}. Hierarchical structure has also been noted in Wikipedia topic networks, in which concepts tend to maintain several connections to the foundational concepts used in each article \cite{Fang2011WikipediaAD}. A hierarchical structure of mathematics knowledge is intuitive, particularly within a delimited area such as linear algebra: a set of foundational concepts, such as matrix, vector, and linearity, are used to motivate and develop the rest of the topics within the field, which, for the most part, all presuppose the concepts in the core. Naturally, this hierarchy will not be a simple dichotomy (core-periphery), but the subsidiary concepts should themselves fall hierarchically into different groups, which may differ across texts due to author interests and publisher goals.

The observed growth dynamics of core-periphery structure offers a coherent expositional model. Given that the set of core concepts are highly related, and thus plausibly represent the concepts providing the foundation for linear algebra, it seems reasonable to introduce these concepts early and to introduce periphery concepts, which presuppose the core, later. The importance of giving sufficient foundational context and prior knowledge in exposition is well-appreciated \cite{problembased}. Further, the edge dynamics we observe are consistent with an expositional model in which topics are procedurally related to each other; the core concepts are introduced first, and used to introduce, at each point in the exposition, the communities that are being focused on; furthermore, subsidiary concepts that have already been introduced may then be used to give context for and develop further, separate subsidiary communities. Such an expository approach, in which connections are consistently introduced between that which has been learned, and that which is to be learned, has been demonstrated as useful in teaching mathematics proofs \cite{letusdefine}. An extension of this motivating expository style is one that incorporates the historical context \cite{fried_jahnke_2015}. In future, it could prove fruitful to compare the expository structure of mathematics texts and the historical development of the results included in those texts.

It is worth noting that, while recent efforts address the dynamics of core-periphery \cite{csermely,Verma2016EmergenceOC} and community \cite{bassettcommunity,dynamiccommunity} structures in networks, comparatively little work has addressed the growth and emergence of these structures over time. The perspective of our work is therefore important; we consider there to be some \emph{a priori} structure of mathematical knowledge, unbeknownst to us, which each author seeks to convey. Thus, rather than examining the evolution of meso-scale features in the networks, we instead focus on how the eventual features, which we take to represent those present in the latent structure of the knowledge, are created throughout exposition. Such methods dealing with the emergence of meso-scale features could prove useful in studies of learning. For example, how do semantic networks of students' knowledge evolve as students are taught? Can that evolution be formally predicted by a generative network model built from the textbook used in their class?

\subsection*{Knowledge gaps in the exposition of mathematics texts}

Using the tool of persistent homology, we examined the growth and persistence of knowledge gaps (colloquially), or topological cavities (formally), in the semantic networks of linear algebra texts. While this tool has been applied to other types of text and knowledge, including Shakespeare's plays \cite{rieck}, natural language \cite{zhu}, discourse \cite{savle-etal-2019-topological}, and collections of mathematics papers \cite{salnikov_co-occurrence_2018}, little is known about how knowledge gaps within a single expositional text or growing semantic network may impact how that text or knowledge structure might be received or understood. Our hypothesis, motivated by the idea that a topologically complex structure with many gaps in knowledge might be more difficult to learn, was that effective exposition likely seeks to produce a smaller number of knowledge gaps, as the creation of a great number of topological cavities could prove confusing to a reader. Still, leaving a few gaps throughout exposition can add intrigue to the subject, piquing the reader's curiosity to make connections themselves \cite{loewenstein_psychology_1994}. 

In the context of this discussion, it is interesting to contrast the features of a process that humans have arguably optimized for explicit learning with the features of a process that nature has arguably optimized for implicit learning \cite{seger1994implicit}. As a token of the former, we consider textbook writing; as a token of the latter, we consider language acquisition in children \cite{steyvers,hills2009longitudinal}. Evidence suggests that knowledge gaps, detected as topological cavities, are a robust feature of language acquisition in toddlers and their prevalence is unaffected by maternal education or by the order in which words are learned \cite{sizemore}. One could speculate that this observed homogeneity in the early semantic feature network learning supports robust language acquisition, ensuring that children who are exposed to different sets of words at different times are still able to reach adult language proficiency. In contrast, when constructing an exposition for a textbook whose sole purpose is to take a set of students from naivety to sophistication in the same place and at the same time, such robustness is not needed and instead consistency, thoroughness, and comprehensiveness is required. The relative paucity of knowledge gaps in the textbooks we study here would be consistent with these distinctions in goals and environment. It could prove useful in the future to more generically assess the robustness of growing networks to the order of node introduction \cite{blevins-reorderability}, particularly to assess differences between implicit or explicit learning processes.

Notably, we found that most cavities that were introduced were eliminated before the end of each text. We observed that, while multiple connected components were introduced, all were eventually -- and usually quite quickly -- connected into a single connected component, suggesting that the expositional order of introduction of edges throughout the text minimizes the extent to which cavities are formed. Remarkably, though, the order of the expositions -- that is, the extent to which cycles were not introduced and did not persist -- did not appear to be maximal. That is, the node-ordered filtration null model exhibited significantly sparser persistent homology than we observed in the texts (Supplementary Fig.~\ref{fig:oaatbarcodedens}). This observation suggests a tradeoff between topological order and apparent learnability; specifically, while such neatly-ordered expositions might minimize the extent to which knowledge gaps are created and persist, it is likely in the best interest of readable and enjoyable exposition to not follow this purely structural ordering -- that is, to properly motivate concepts, give relationships where they might seem natural and useful, and make the text generally more readable.

Our correlation analysis of the barcode densities suggests some interesting directions for further study of the potential relationship between persistent homology of a growing semantic network and effective learnability. Specifically, while our study did not deal explicitly with differential learnability of texts or in how knowledge gaps might affect the learning process, we did observe several interesting relationships between the 0- and 2-dimensional barcode densities and textbook ratings. While these results are preliminary, they suggest that an interesting avenue for further study would be to examine the topology of growing semantic networks in the classroom setting. In particular, one could consider multiple networks: the network of the textbook being used, providing the ``latent space'' of the knowledge and the relationships between concepts; the teacher's network, as provided in class to the students through lessons; and finally, the students' networks, as they develop over time while the students learn the material. An analysis of the developmental and topological relationships between all three of these classes of semantic networks could yield interesting results in how knowledge structures are transferred from teacher and book to student, and could provide useful insight to effective structuring and expositional presentation of knowledge in a textbook format.

\subsection*{Methodological considerations}

There are certain limitations inherent in our work that should be considered for future study. First, our text extraction methodology imperfectly converted PDFs to plaintext, leaving significant textual noise and artifacts of embedded math which required subsequent automated removal, and the remnants of which prevented perfect concept extraction and sentence-level co-occurrence calculation. Because textbook PDFs are easier to access than textbook source material, we spent significant time developing our text extraction approach to account for these circumstances so that our methodology could be widely applicable. However, future work could utilize the LaTeX source for textbooks in order to reduce noise. Second, the problem of concept extraction is ill-posed due to the subjectivity of the notion of ``concept''. We examined a number of supervised and unsupervised keyphrase extraction algorithms, and our modified RAKE algorithm performed best in comparison to our intuitive expectations for linear algebra contexts. However, future work will be necessary to better understand (a) how to determine how many concepts should be extracted from a text, (b) what should comprise a ``concept'' in a semantic network, and to (c) examine hierarchically structured semantic networks to incorporate the subjectivity of concepts into the network structure, so that high-level concepts are distinguished from those which are lower-level. Third, our network and filtration construction methodology is only one of many possible methodologies; as we chose to use co-occurrence to construct the networks, they are undirected and lack edge labels detailing the nature of each relationship. Fourth, the application of a clique complex to infer knowledge gaps in a growing network is one of many choices, and it assumes that any fully-connected $(k+1)$-cliques should, in fact, reflect a filled $k$-simplex of knowledge. However, a possible alternative could be to only add a $k$-simplex when such higher-order relationships are observed simultaneously, such as when three words co-occur in the same sentence. Finally, further research in a classroom setting should be able to provide insight into what types of knowledge gaps might have an effect on student learning, thus providing an answer as to how persistent homology should be computed on growing semantic networks.

\subsection*{Future directions}

A clear open area for future work lies in understanding tradeoffs in ordered network structure. Here, we find four separate instances in which semantic networks of linear algebra textbooks appear to balance competing constraints. First, while core-ness and modularity are higher than expected in a continuous configuration null model, they are notably lower than expected in a random index null model. Second, while core nodes tend to be added more quickly than periphery nodes, the difference in speed is more stark in the random index model. Third, while some texts add core nodes faster than expected in the random sentence order null model, some texts add core nodes more slowly, suggesting that each text opts for a different expositional style. Fourth, while the barcodes of the empirical networks are relatively sparse compared to the random edge model, they still exhibit more persistent cycles than the most ordered model, the node-ordered filtration. Collectively, these results suggest that effective and useful exposition, while structured in nature, is not as strongly structured as it could be. It may be effective to purposefully introduce some gaps in knowledge by withholding topics to support productive failure \cite{kapur} or provide detailed motivation to stimulate curiosity \cite{loewenstein_psychology_1994, wade_role_2019}. Of course it is also possible that our observations reflect the nature of the structure of mathematics: perhaps mathematics simply does not have as strongly-ordered a structure as we might observe in our null models. Future efforts could seek to better understand this tradeoff and its potential causes.

\section*{Materials and methods}
All tools and methods developed for use in this work are designed to be broadly applicable to any expositional text. We thus provide Python code for the extraction and analysis of semantic networks at \textit{https://github.com/nhchristianson/Math-text-semantic-networks}. Further details and considerations for the methods used can be found in the Supplementary Methods.

\subsection*{Data collection and preprocessing}
We collected a diverse set of ten linear algebra textbooks in PDF format, ranging in focus from theory to application (see Supplementary Methods for more details). We converted the PDF files to plaintext with the tool at \textit{https://pdftotext.com}, and manually cleaned each text to isolate the main chapters, discarding introductory or appendix sections. We then converted the text to unicode KD normal form, replaced hyphens with spaces, and used spaCy \cite{spacy2} to lemmatize all words in each text, which reduces inflected words to their dictionary form. We then used the Python Natural Language Toolkit (NLTK, Version 3.3 \cite{BirdKleinLoper09}) to tokenize the text into sentences and their component words, replacing any word containing numerical characters with the character ``\#''. We then removed all words not comprised solely of letters and ``\#'' and made the remaining words lowercase. Due to the presence of embedded mathematics in the textbooks which is imperfectly handled by the PDF-to-text conversion process, we implemented a final measure in an attempt to clean the text of artifacts such as the remnants of variables and equations. In particular, we first created a ``stop list'' -- that is, a list of common ``stop words'' in English -- by removing all single-letter words from the Ranks NL Long stop word list (\textit{https://www.ranks.nl/stopwords}), since such single character words likely represent variables within the text. Then, we applied a series of rules to determine whether any word token was sufficiently variable-like to be converted to a ``VAR'' variable placeholder: any ``\#'' placeholder was kept as is; then, any word without vowels was converted to ``VAR''; then, any word of length at most two not present on our stop list was converted to ``VAR''; and finally, any word of length 3 or 4 that did not spell check using the Enchant \cite{enchant} spell-checker was converted to ``VAR''. 

\subsection*{Concept extraction}
The linguistics and natural language processing literature provide a number of canonical statistical metrics for determining the significance of $n$\textit{-grams}, or phrases comprised of $n$ words, within text \cite{manningschutze}. After testing multiple supervised and unsupervised keyphrase extraction methodologies, we chose to use an unsupervised method based on the rapid automatic keyword extraction (RAKE) algorithm \cite{RAKE} to extract concepts from our texts. RAKE works as follows: 
\begin{enumerate}
	\item A provided set of stop words, phrase delimiters, and word delimiters are used to divide the document into a set of candidate keyphrases and their comprising keywords.
	\item The frequency of keywords and their co-occurrence in different keyphrases is calculated, forming a co-occurrence graph.
	\item The candidate keyphrases are ranked by a scoring function score$(k)$, which typically ranks candidate keyphrases by certain properties of their comprising keywords.
	\item Some threshold $n$ is chosen, and the top $n$ ranked candidate keyphrases are kept as the extracted keyphrases.
\end{enumerate}
In RAKE, the scoring function for a candidate keyphrase $k$ is typically taken to be
$$\text{score}_{\text{RAKE}}(k) = \sum_i \frac{\text{deg}(k_i)}{\text{freq}(k_i)},$$
where deg$(k_i)$ and freq$(k_i)$ are the degree and frequency, respectively, of the $i$th keyword comprising the phrase $k$ in the co-occurrence graph RAKE constructs. As such, RAKE poses that significant keyphrases are those whose component words co-occur with many other words, but do not occur very frequently. Because we wish to ensure that the scores of more plausibly mathematical words are high, we modify this keyphrase scoring function to incorporate the term frequency-inverse document frequency ranking method \cite{tfidf}, including an additional term to account for a given keyphrase's frequency in an external corpus. Specifically, we specify our phrase scoring function as
$$\text{score}(k) = \frac{\text{score}_{\text{RAKE}}(k)}{1 + \text{brown}(k)},$$
where brown$(k)$ is the number of times that the whole keyphrase occurs in the Brown corpus \cite{kucera} with the ``Learned'' category (comprised of scientific and other academic texts) removed. As such, we aim to penalize phrases that occur very frequently in non-mathematical text, as such words will likely not be mathematically meaningful. We add 1 to the brown$(k)$ term in the denominator since not all phrases RAKE extracts occur in the Brown corpus. Details on our specific implementation of the modified RAKE algorithm can be found in the Supplementary Methods.

\subsection*{Network construction}

We construct each text's semantic network of concepts by calculating the co-occurrence of concepts in each text on a sentence level. That is, we deem two concepts in each text's index set to co-occur, and thus be related, if they occur in the same sentence at some point in the text. We also assign to each edge between concepts an integer weight indicating the number of sentences in which the two concepts co-occur. This data yields an undirected weighted graph $G = (V, E)$, where each node $v \in V$ is a concept and each edge $(v_1, v_2) = e \in E$ represents a semantic relationship between concepts with an associated positive integer weight $w(e) \in \mathbb{Z}^+$ denoting the number of sentences in which the two concepts co-occur.

We are not merely interested in the total semantic network of each textbook, but in the development of the semantic networks over the course of exposition. Thus, for each text, we keep track of the first sentence in which each concept and each relationship -- equivalently, each node and each edge -- is introduced. If a text has $N$ sentences, our methodology of extracting growing semantic networks yields a sequence of $N$ graphs $G_1 \to \cdots \to G_N$, where the $k$th graph $G_k$ includes all nodes and edges which have been introduced prior to or during the $k$th sentence of the text. In the context of algebraic topology which we employ throughout this study, such a sequence of nested objects is called a filtration. We call this sequence of graphs the \textit{expositional filtration} of a text. In considering this filtration, we consider the binarized graphs; that is, we disregard edge weight data during the exposition, only considering edge weight data for the final semantic network, which we call the \textit{total network}.

\subsection*{Meso-scale network structure}

Complex networks often exhibit meso-scale or global characteristics of structural order. Certain networks exhibit \textit{community structure}, in which densely connected \textit{communities} of nodes exhibit sparse or weak inter-community connections \cite{Newman8577}. In the context of semantic networks, such densely connected communities may represent strongly related concepts that indicate the existence of some higher-order enveloping concept or umbrella term. Another type of meso-scale structure which may be exhibited is \textit{core-periphery} structure, which is characterized by a densely connected set of \textit{core} nodes and a set of \textit{periphery} nodes which are sparsely connected amongst themselves, but are strongly connected to the core \cite{Borgatti2000ModelsOC}. Such an organization of semantic networks is plausible in the context of mathematics, in which many different ideas may be developed from a smaller set of highly related concepts.

To detect community and core-periphery structure in the networks, we used the Brain Connectivity Toolbox for Python, version 0.5.0, which is based on the MATLAB Brain Connectivity Toolbox (BCT) \cite{RUBINOV20101059}. To evaluate the presence of a core-periphery structure, we seek to assign a network's nodes to either the core or the periphery group so as to maximize the \textit{core-ness} quality function \cite{Rubinov10032}:
$$Q_C = \frac{1}{v_C}\left(\sum_{i, j \in C_c}(w_{ij} - \gamma_C\bar{w}) - \sum_{i, j \in C_p}(w_{ij} - \gamma_C\bar{w})\right),$$
where $C_c$ and $C_p$ are the sets of nodes in the core and periphery, respectively, $w_{ij}$ is the weight of the edge from node $i$ to node $j$ (which will be 0 if the nodes are not connected by an edge), $\bar{w}$ is the average of all edge weights, where nonexistent edges with ``zero weight'' are also included in the average, $\gamma_C$ is a resolution which controls the size of the core, which we set to 1, and $v_C$ is a normalization constant. In effect, in maximizing core-ness we seek to maximize the number and weight of intra-core connections, while minimizing the number and weight of intra-periphery connections.

To evaluate the presence of community structure in the networks, we use a Louvain-like locally greedy algorithm \cite{1742-5468-2008-10-P10008} to optimize the modularity quality function:
$$Q_M = \frac{1}{v_M} \sum_{i, j \in C} \left(w_{ij} - \gamma_M \frac{s_is_j}{v_M}\right)\delta_{ij},$$
where $C$ is the set of network nodes, $w_{ij}$ is the weight of the connection from node $i$ to node $j$, $s_i$ and $s_j$ are the summed weights of edges connected to node $i$ and node $j$, respectively, $\gamma_M$ is a resolution parameter controlling the size of communities which we set to 1, $v_M$ is a normalization constant, and $\delta_{ij}$ is the Kronecker delta function, which is 1 when node $i$ and node $j$ are in the same community and is 0 otherwise \cite{Rubinov10032}. In effect, modularity maximization seeks to maximize the strength and number of connections within communities, yielding a partition of the network nodes into a set of densely connected communities with few inter-community connections.

\subsection*{Persistent homology}

Beyond characterization of the local and meso-scale attributes of the total semantic network of the texts, we furthermore seek to evaluate structural and topological characteristics of the semantic networks as they are built over the course of the entire text. In particular, we study the extent to which ``knowledge gaps'' are created and persist in semantic networks throughout a text's exposition. To this end, we use a method with roots in the mathematics of algebraic topology called \textit{persistent homology} which, in short, evaluates the creation and lifespan of topological ``holes'' in data over time, or in this case, over the course of exposition, thus allowing us to characterize and evaluate the presence of these gaps in knowledge. Here we give a brief, intuitive overview of how we calculate persistent homology for our expositional semantic networks; the particularly interested reader may refer to Refs. \cite{carlsson, zomorodian, Edelsbrunner2013PersistentHT} for a rigorous overview of persistent homology and its computation for data analysis, as well as Refs. \cite{horak, petri2013, stolz2017, Otter2017, Bampasidou2014ModelingCW} for example uses of persistent homology in the context of complex networks.

Recall that a text's semantic network at a certain point in the exposition (a particular graph in the expositional filtration) is an undirected graph, where connections between nodes indicate that the concepts represented by those nodes have already co-occurred in a sentence. Given a binary undirected graph $G = (V, E)$, we may construct an object called the \textit{clique complex} $X(G)$, which, for every natural number $k$, assigns to every all-to-all connected subgraph of $G$ on $(k+1)$ vertices (also known as a $(k+1)$-clique) a $k$-\textit{simplex}, which may geometrically be represented as the convex hull of $(k+1)$ affinely independent points. For example, a 0-simplex is simply a single node, a 1-simplex is an edge, a 2-simplex is a filled-in triangle, and a 3-simplex is a filled-in tetrahedron. Intuitively speaking, this clique complex $X(G)$ is a ``filled-in'' version of the graph $G$, where, for each $k$, we choose a distinct color and then color in all $(k+1)$-cliques in $G$ to form $k$-simplices. Then, classical \textit{homology} intuitively describes, for each $k$, how many topological ``holes'' of dimension $k$ are in the complex, or how many regions are enclosed by the $k$th color, but are themselves not colored as such. In other words, homology detects \textit{cycles}\footnote{More precisely, homology finds equivalence classes of cycles, but we refer to an equivalence class as a cycle for simplicity.} of $k$-simplices that surround a void. For example, a 1-cycle reflects a conventional cycle in a graph, just like the hole in a circle; and a 2-cycle reflects a cavity, like the hole in the center of a sphere. A 0-cycle is intuitively slightly different, in that 0-cycles refer to connected components of the graph, so that having more than one 0-cycle tells us that multiple disconnected components exist. In our work, we restrict our focus to these first three dimensions, since these are the most geometrically intuitive. These cavities or holes are exactly the knowledge gaps we seek in the semantic networks, as they indicate some closed cycle of $(k+1)$-order connections between concepts surrounding a region of lesser connectivity. 

A useful extension of homology enables us both to count the number of holes present in the semantic network at each step in a text's exposition, as well as to keep track of which topological cavities are created and destroyed at each step. Specifically, \textit{persistent homology} allows for the computation of the homology for the sequence of clique complexes of our expositional filtration $X(G_1) \to \cdots \to X(G_N)$; this tool not only keeps track of the number of cavities of each dimension present at each expositional step, but it also tracks the \textit{persistence} of each individual cavity over the course of the exposition, so we may identify individual knowledge gaps, when they were created, how long they persist, and when they are extinguished. Rigorously, the $k$th persistent homology of a graph filtration yields a (multi-)set of intervals called the \textit{barcode}:
$$\{[b_1, d_1), \ldots, [b_m, d_m)\},$$
where $b_i$ indicates the time of birth of the $i$th $k$-dimensional cavity, and $d_i$ indicates the time of death of that cavity (which may be $\infty$ if the cavity is still present in the total network, i.e., it never dies). Thus, the number of intervals, as well as their length (the difference between their death and birth times), indicate the number and persistence, respectively, of topological cavities during exposition. 

Once we have computed the persistent homology of a text's expositional filtration for a given $k$, we may use several characteristics of the resultant persistence intervals to examine various aspects of the persistence of knowledge gaps in the semantic networks. In particular, we consider two metrics: first, we examine the value of $m$, which gives the total number of $k$-cavities which were created, at some point, over the course of the exposition. Secondly, we define a metric similar to one presented in Ref. \cite{adcock}, which we refer to as the \textit{normalized average cycle lifetime} of dimension $k$:
$$D_k = \frac{1}{mN}\sum_{i=1}^m (d_i - b_i),$$
where $N$ is the number of steps in the filtration, and $d_i$ is the time of death of the $i$th $k$-cavity, unless $d_i = \infty$, in which case we set $d_i = N+1$, to distinguish these infinitely-persisting cavities from those that die at step $N$. Intuitively, this metric describes the extent to which an expositional filtration has cycles which are persistent; it is normalized by $N$, the length of the filtration, and $m$, the number of $k$-cycles introduced throughout the filtration, so as to be comparable across texts which might have different filtration lengths or total numbers of cycles introduced. The goal is to allow a formal comparison of how persistent $k$-cycles tend to be in different texts.

In our work, we use Ripser.py \cite{ctralie2018ripser} due to its speed and efficiency for the computation of persistent homology for the empirical networks and the null models. 

\subsection*{Null models}

In order to determine to what extent the results we obtain for meso-scale structure and topological dynamics in the semantic networks are significant, we employ two categories of null models: data-level null models, which randomize on the scale of the underlying text and index list, from which we may then extract semantic networks and expositional filtrations; and projected network-level null models, which randomize on the scale of the networks we extract for each text. Furthermore, while some of these models are particularly suited as null models for the structural metrics on the total network since they yield a single, weighted network, others are more suitable as null models for the growing dynamics of the semantic networks, as they provide a null expositional filtration. For each null model, our null ensemble is comprised of 100 random instantiations; we present the resulting null distributions of metrics alongside the data for our actual networks in our results. Here we summarize the null models and their uses; more detailed descriptions of the models can be found in the Supplementary Methods.

\begin{enumerate}[(a)]
	\item Random index: Expositional filtration of randomly chosen words in each text, to serve as a semantic network on random ``concepts''. Acts as a null model for the total network and empirical filtration.
	\item Random sentence order: Expositional filtration of original index set with randomly-shuffled sentences, keeping sub-sentence structure while randomizing exposition on the sentence scale. Acts as a null model for the empirical filtration.
	\item Continuous configuration: Rewired network preserving node degree and strength. Acts as a null model for the total network.
	\item Random edge: Random reordering of edge introduction from the empirical filtration, mimicking totally random exposition. Acts as a null model for the empirical filtration.
	\item Node-ordered: Adds each node and all its edges in order of node introduction in the exposition, mimicking exposition that connects each concept to all previously-related concepts. Acts as a null model for the empirical filtration.
\end{enumerate}

\section*{Acknowledgments} 
We are grateful to David Lydon-Staley, Dale Zhou, Alec Helm, and Shubhankar Patankar for their generous advice on early versions of this manuscript. D.S.B., N.H.C., and A.S.B. acknowledge support from the John D. and Catherine T. MacArthur Foundation, the Alfred P. Sloan Foundation, the ISI Foundation, the Paul Allen Foundation, the Army Research Laboratory (W911NF-10-2-0022), the Army Research Office (Bassett-W911NF-14-1-0679, Grafton-W911NF-16-1-0474, DCIST- W911NF-17-2-0181), the Office of Naval Research, the National Institute of Mental Health (2-R01-DC-009209-11, R01-MH112847, R01-MH107235, R21-M MH-106799), the National Institute of Child Health and Human Development (1R01HD086888-01), National Institute of Neurological Disorders and Stroke (R01 NS099348), and the National Science Foundation (BCS-1441502, BCS-1430087, NSF PHY-1554488 and BCS-1631550).

\clearpage
\bibliography{references}

\clearpage

\beginsupplement

\section*{Summary of Supplementary Material}
In this supplementary document, we provide supplemental methods, followed by supplemental results. We conclude with additional discussion relevant to the findings in both the main and supplemental texts.

\section*{Supplementary Methods}

\subsection*{Textbooks used}
The ten textbooks used in our study \cite{axler,bretscher,edwards,greub,hefferson,lang,petersen,robbiano,strang,treil} have publication dates ranging from 1967 to 2018. The set also includes two texts that were translated from a different language, and two texts that are made available online for free use.

\subsection*{Considerations for concept extraction}

In order to construct a semantic network, it is first necessary to choose which concepts should comprise the nodes of that network. Much previous work has considered all or most of the individual words in a text as the network nodes \cite{lucychai,PEREIRA20111192}; we avoid this assumption so that we may consider, further than individual words, higher-level concepts that may be presented in multi-word phrases. Another choice of nodes could be the topics present in the index of a text, if an index is included. We also choose not to use this method, as we seek to determine and extract the concepts from the text's exposition via some more intrinsic metric of conceptual significance. This choice was motivated by an interest in examining the semantic networks of concepts that the text poses as significant, rather than simply those of concepts which the author deems significant. Thus, via this paradigm of intrinsic conceptual significance, we aim to emulate human readers in their assessment of the significance of concepts. In choosing a methodology of extracting concepts from the texts for use as the networks' nodes, we sought to find a method that would maximize the number of extracted mathematical concepts while minimizing the number of extracted words and phrases that are not mathematics related. We also sought a method that would be extensible to domains of knowledge and exposition aside from mathematics, so that our whole methodology can be extended to the analysis of general textual exposition. These considerations led to our development of the modified RAKE algorithm.

\subsection*{Implementation details for our concept extraction methodology}

In our code, we use the python-rake implementation of RAKE (\textit{https://github.com/fabianvf/python-rake}); as a stop word list, we use the modified Ranks NL Long stop word list we discuss in the main text, from which we remove the word ``value'',which plays an important role in linear algebra phrases such as ``singular value decomposition''. We also add to this stop list our placeholder words ``\#'', ``VAR'', and ``-pron-'' (the pronoun placeholder output by the spaCy lemmatizer), as well as certain words used extensively in mathematics exposition that do not convey mathematical content, in an effort to ensure that our keyphrases might better reflect a set of meaningful mathematical concepts (see Supplementary Table~\ref{table:stop_words}). We also prune the candidate pool by specifying that keywords must be comprised of at least 3 characters and must occur at least 5 times within the text, and keyphrases can be no more than 4 words long. Given these specifications, RAKE generates a set of candidate keyphrases and their associated scores, which we modify with the addition of our extra Brown frequency term. We then clean the candidate keyphrases by removing any of the numerical, variable, or pronoun placeholders; after this cleaning, if there are any duplicate candidates, we give the keyphrase in question the highest score from all duplicates. We choose to keep the top-scoring half of candidate keyphrases, since this threshold appears to include most phrases one might expect to represent significant linear algebra concepts in each text; thus we take the top half of scored keyphrases to be the concept set for each text, which we refer to as the \textit{index list}. This threshold of one-half is similar to thresholds used in other work, such as the threshold of one-third in RAKE \cite{RAKE} and Textrank \cite{textrank}. However, no choice of threshold will perfectly include all relevant concepts and omit irrelevant words.

\begin{table}
\centering
\begin{tabular}{|c|c|c|c|c|c|}
\hline
examples & counterexample & text & texts & undergraduate & chapter \\\hline
definition & notation & proof & exercise & result & \\\hline
\end{tabular}
\caption{Common words in mathematical exposition that we add to the stop word list for concept extraction}
\label{table:stop_words}
\end{table}

\subsection*{Considerations for network construction}

\begin{figure}
    \centering
    \begin{subfigure}[c]{0.45\textwidth}
        \centering
        \[\begin{tikzcd}
         \text{square matrix} \arrow{rr}{\text{has a}}\arrow{dd}{\text{may be an}}&&\text{determinant}\arrow{dd}{\text{is the product of}} \\
         && \\
         \text{isomorphism}\arrow{rr}{\text{has only nonzero}}&&\text{eigenvalues}
        \end{tikzcd}\]
        \caption{Unfilled knowledge gap in the network}
    \end{subfigure}
    \begin{subfigure}[c]{0.45\textwidth}
        \centering
    \[\begin{tikzcd}
         \text{square matrix} \arrow{rr}{\text{has a}}\arrow{dd}{\text{may be an}}&&\text{determinant}\arrow{dd}{\text{is the product of}} \\
         && \\
         \text{isomorphism}\arrow{rr}{\text{has only nonzero}}\arrow[rruu, "\text{has nonzero}", pos=0.7]&&\text{eigenvalues}
        \end{tikzcd}\]
        \caption{Knowledge gap is filled}
    \end{subfigure}
    \caption{A simple example of a semantic network comprised of linear algebra concepts. \emph{(a)} The lack of connection between ``square matrix'' and ``eigenvalues'' or between ``isomorphism'' and ``determinant'' indicates the presence of a knowledge gap. \emph{(b)} The knowledge gap is extinguished by the addition of the relationship between ``isomorphism'' and ``determinant,'' thus ensuring that all concepts' neighbors are also neighbors themselves.}
    \label{fig:semantic_network_ex}
\end{figure}
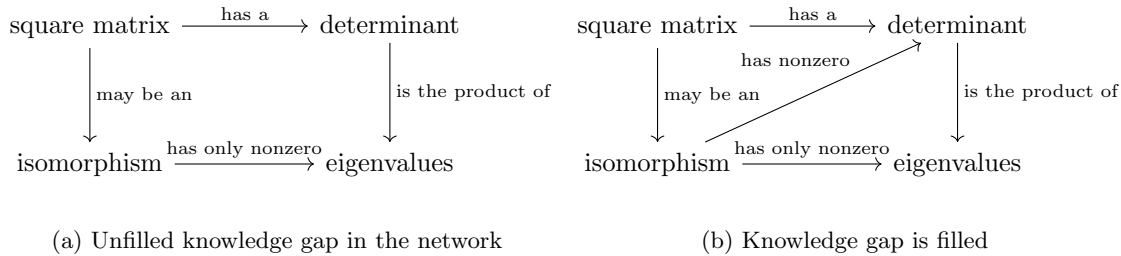

Once we have determined a set of concepts to use as the nodes of a text's semantic network, we then wish to form the semantic network of those concepts and their relationships, as provided by the text's exposition. Certain approaches to semantic network construction seek to determine not only whether two entities are related, but also the semantic nature of the relationship between the entities in question. Fig.~\ref{fig:semantic_network_ex} gives an example of such an annotated semantic network, in which each relation has a meaningful label. Such semantic parsing techniques to generate semantic networks have been applied to scientific texts in several cases \cite{parsing_pharmacogenomic, semeval}, but they generally require involved syntactic parsing rules or data annotation. We did not use these approaches, as the messy nature of the text-converted mathematics textbooks -- with embedded variables, formulas, and symbols sometimes interjecting sentences -- likely would have interfered with effective inference of semantic relationships. Instead, we use a method of extracting concept relationships that is more resistant to such noise: co-occurrence frequency \cite{cantbeatfrequency}. Co-occurrence is a notion specifying the degree to which words or phrases tend to occur nearby each other in either a text or a set of texts. Statistical metrics based on co-occurrence have been studied extensively in the field of computational linguistics as a measure of the semantic relatedness of words or phrases \cite{Schutze93wordspace, Mcdonald98modellingfunctional, Bullinaria2007}. Because we are interested in relationships between concepts which are not purely linguistic in nature, and since many of our extracted concepts are multiple-word phrases, we choose to calculate co-occurrence on the sentence level; this level of granularity will also ensure that phrases in the same sentence, yet separated by a string of math variables, will be inferred to be related. 

\subsection*{Null models}

Here, we describe in more detail the construction and role of each null model we employ in our work. We begin with the data-level null models: for both the total network and the expositional filtration, we wish to determine the extent to which our results might simply be reflective of the topology one would expect from the growing ``semantic network'' generated by computing the co-occurrence of a random set of words in our texts. To this end, we employ a \textit{random index} null model, in which we select a random set of index terms of equal size to the original index list, drawn without replacement from the set of words comprising each text (not including the augmented stop word list we used for RAKE extraction). We use this random index list as our set of ``concepts'' for calculating each text's co-occurrence, yielding both a final weighted network, as well as an expositional filtration, allowing this null to be used both in the comparison of meso-scale structure and development, as well as of persistent homology. Note, however, that we may interpret the random index null model in a different way: that is, since the random index set excludes any stop words, it must be comprised of meaningful words. Thus, the random index model can be viewed as conveying a semantic network -- not the network that the book intends to convey, but a semantic network nonetheless that may very well include some mathematically meaningful concepts.

We further seek to establish the extent to which our results on topological development of the networks are dependent on the order in which relationships are introduced within the texts. We therefore employ a \textit{random sentence order} null model, in which for each text, we randomly permute that text's sentences, and use the original set of index terms to calculate co-occurrence. This null model yields the same total network, since the index set is the same and the same sentences are present, and thus the same sentence-level co-occurrences will occur; however, the filtration it yields will differ in the order of edge introduction, thus enabling us to study how the meso-scale and topological development of the network differs based on differing sentence order.

The remainder of our null models are projected network-level nulls. To evaluate the extent to which the results we observe for the core-periphery and community structure of the empirical networks would be expected from a random network with a similar joint distribution of node degrees and weights, we use the \textit{continuous configuration model} \cite{contconfig}. This model is an extension of the configuration model for random graph generation, and seeks to preserve the expected degree of each node, as well as the expected strength of the node, where a node's strength is the sum of the weights of the edges it participates in. Specifically, if $d_u$ and $s_u$ give the degree and strength, respectively, of a node $u$, and $d_T$ and $s_T$ are the sum of all node degrees and strengths, respectively, then given some graph with node set $[n]$, for any two nodes $u, v \in [n]$, we define $d_{uv} = \frac{d_ud_v}{d_T}$ and $s_{uv} = \frac{s_us_v}{s_T}$, as well as $\{P_{uv}\}$ as some family of probability distributions with mean one. Then to generate a graph using the continuous configuration model, we iterate through all possible pairs of nodes $u, v$, introducing an edge between $u$ and $v$ with probability $d_{uv}$; if an edge is introduced, then the edge is given weight $w_{uv} = \frac{s_{uv}}{d_{uv}}\xi_{uv}$ where the normalized weight random variable $\xi_{uv} \sim P_{uv}$. For the sake of simplicity, we assume that all distributions $P_{uv}$ are identical, so that all $\xi_{uv} \overset{iid}{\sim} P$; we discuss our fitting of the distribution $P$ for each network in the supplement.

To examine how our results on persistent homology differ from a model of exposition in which connections are drawn completely at random -- that is, with a filtration of the empirical total network that adds edges randomly -- we employ the \textit{random edge} null model. In this model, edges present in the empirical total network are introduced in a random order, and nodes are introduced immediately preceding their first inclusion in an added edge. Next, to determine how our persistent homology results differ from a model of exposition in which concepts are iteratively introduced and connected to all already-introduced concepts, we examine a \textit{node-ordered filtration} \cite{sizemore, blevins-reorderability}. In this null model, nodes are added by order of introduction in the text; if multiple nodes were originally added in a single sentence, then those nodes will be added to the node-ordered model in a random order. After each node is added to the null, all edges between it and previously-added nodes that are present in the total network are added in a random order. 

Finally, we must consider a caveat for the filtration null models: in particular, while the original expositional filtration, the random index null, and the random sentence order null have some intrinsic sense of ``time'' of introduction due to the presence of the sentence structure of the text, the latter two null models do not, as they introduce nodes and edges one at a time. As such, in order to meaningfully compare persistence barcodes amongst all these models, we must ``unfurl'' the expositional filtrations of the real network and the random index networks. To this end, we introduce the one-at-a-time (OAAT) filtration process; this methodology takes a filtration in which multiple nodes and edges might be introduced in single sentences, such as the expositional filtration of a text, and transforms it so that only a single node or edge is added at each step in the filtration. Specifically, for each sentence, the OAAT process examines what nodes and edges are added to the network in that sentence; if multiple nodes are added, then they are added first, one at a time, in a random order; then edges are added, one at a time, in a random order. For our empirical expositional network, we compute 100 instantiations of this OAAT filtration in order to account for stochasticity in the random ordering (we do not do this for each random index or sentence order filtrations, since we already compute 100 distinct such graphs). With this method, we may examine the topological development that occurs not just over the course of the text with a sentence-level granularity, but also on a sub-sentence scale.

There are certain tradeoffs we make in using the OAAT filtration for our expositional filtrations. In particular, we lose the direct relationship of cavity persistence length to ``time'', or sentence duration throughout the text, since we instead simply introduce one node or edge at each ``timestep'' in the OAAT filtration. However, long cycles should still tend to be long, under the assumption that there is relatively consistent introduction of nodes and edges throughout the texts. Furthermore, this ``unfurling'' of the expositional filtration gives us the ability to do a tête-à-tête comparison of our latter two null models to the expositional filtrations. These two nulls have no built-in notion of time, and introduce a single node or edge at each step of their filtration; as such, putting our expositional filtrations on equal footing makes the qualitative and quantitative comparison of the persistent homologies of these filtrations more direct.

\section*{Supplementary Results}

\subsection*{Estimating the normalized weight distributions for the continuous configuration model}

The parametrization of the continuous configuration null model for weighted undirected graphs rests upon the choice of a family of probability distributions $P_{uv}$ that specifies the distribution of the possible ``normed weight'' values for each edge connecting nodes $u$ and $v$ in a network's node set. Specifically, where $d_u$ and $s_u$ are the degree and strength, respectively, of a node $u$, and $d_T$ and $s_T$ are the sum of degrees and strengths respectively over all nodes in a network, and $d_{uv} = \frac{d_ud_v}{d_T}$ and $s_{uv} = \frac{s_us_v}{s_T}$ give a normalized view of to what extent two nodes are both high (or low) in degree or strength, then the continuous configuration model assumes that the weight of an edge between two nodes $u$ and $v$, if such an edge exists, will be
$$w_{uv} = \frac{s_{uv}}{d_{uv}}\xi_{uv}$$
where $\xi_{uv} \sim P_{uv}$, some probability distribution on what we call the ``normalized weight'' of an edge. In our work, for the sake of simplicity, we make the assumption that all normalized weight distributions are the same distribution $P$. With this assumption, we may choose a parametrization of $P$ and fit this distribution on the empirical normalized weights of all edges in a given network. In particular, if the empirical edge weights are given as $\hat w_{uv}$ for all $u, v$ in the set of edges, then the empirical normalized weights are simply given by $\frac{\hat w_{uv}d_{uv}}{s_{uv}}$.

Once we have the normalized weights, we may choose a parametrization. Because the normalized weights of a network are positive and not restricted to the integers, we attempted maximum likelihood fits of a number of continuous probability distributions with support on the positive real line on each of the networks' normalized weights. Specifically, we focused on long-tailed distributions: the Pareto, Log-normal, L\'{e}vy, Burr, Fisk, Log-gamma, Log-Laplace, and power-law distributions. We also calculated the Kolmogorov-Smirnov (K-S) statistic $D$ of each best-fit distribution in order to determine how well the distribution fit the empirical normalized weight data. Distributions were fit and K-S statistics were calculated in Python with the SciPy library, version 1.1.0 \cite{scipy}. In all networks, the K-S statistic was quite low ($D < 0.025$) with $p$-values all significantly greater than $0.05$, indicating good fit between the empirical and best-fit distributions, or insufficient evidence to reject the null hypothesis that the empirical normalized weight distribution and the best-fit distribution are identical. The best-fits and statistics for each text's network are reported in Table~\ref{table:cont_config_distributions}.

\begin{table}
\centering
\begin{tabular}{|c|c|c|c|}
\hline
Text & Best-fit distribution & K-S statistic & K-S $p$-value \\
\hline\hline
Treil & Burr & 0.0163 & 0.303\\
\hline
Axler & Burr & 0.00997 & 0.795\\
\hline
Edwards & Log-normal & 0.0232 & 0.195\\
\hline
Lang & Log-normal & 0.0140 & 0.687\\
\hline
Petersen & Burr & 0.0165 & 0.174\\
\hline
Robbiano & Fisk & 0.00964 & 0.847\\
\hline
Bretscher & Burr & 0.00870 & 0.758\\
\hline
Greub & Burr & 0.0146 & 0.436\\
\hline
Hefferson & Burr & 0.00696 & 0.910\\
\hline
Strang & Burr & 0.00762 & 0.759\\
\hline
\end{tabular}
\caption{Best-fit distributions and corresponding K-S statistics and $p$-values for the normalized weight distribution of each text.}
\label{table:cont_config_distributions}
\end{table}

\subsection*{Concepts that appear in more than half the semantic networks' cores}

See Table~\ref{table:core_freqs}.
\begin{table}
\centering
\begin{tabular}{|c|c|}
\hline
Concept & Frequency in cores\\
\hline\hline
multiplication & 8 \\\hline
vector space & 7 \\\hline
scalar & 7 \\\hline
vector & 8 \\\hline
inverse & 8 \\\hline
matrix & 9 \\\hline
polynomial & 7 \\\hline
coefficient & 8 \\\hline 
linear transformation & 6 \\\hline
linear & 8 \\\hline
linearly independent & 9\\\hline
diagonal & 9 \\\hline
theorem & 9 \\\hline
projection & 6 \\\hline
orthogonal & 9 \\\hline
invertible & 6 \\\hline
subspace & 9 \\\hline
determinant & 9 \\\hline
diagonal matrix & 6 \\\hline
eigenvalue & 9 \\\hline
eigenvector & 8 \\\hline
orthonormal & 7 \\\hline
orthonormal basis & 6 \\\hline
equation & 7 \\\hline
symmetric & 6 \\\hline
\end{tabular}
\caption{Concepts that occur in more than half of the texts' cores.}
\label{table:core_freqs}
\end{table}

\subsection*{Example concepts in the Axler periphery communities}
See Table~\ref{table:comm_concepts}.

\begin{table}
\centering
\begin{tabular}{|c|c|}
\hline
Community & Example concepts \\
\hline\hline
1 & commutative, associative, dual space, dual map, duality, column rank, row rank \\\hline
3 & finite dimensional subspace, orthogonal, orthogonal complement \\\hline
4 & inverse, additive inverse, additive identity \\\hline
6 & null space, injective, surjective, isomorphism, invertibility, identity map \\\hline
7 & induction hypothesis, division algorithm, factorization \\\hline
8 & linearly dependent, linear combination, orthonormal list, gramschmidt procedure \\\hline
9 & euclidean inner product, dot product, continuous real value[d] function, derivative \\\hline
10 & positive operator, adjoint operator, complexification, complex spectral theorem \\\hline
11 & transpose, permutation, determinant, square matrix \\\hline
\end{tabular}
\caption{Example concepts present within communities in the Axler periphery.}
\label{table:comm_concepts}
\end{table}

\subsection*{Development of the meso-scale core-periphery and community structures}

\begin{figure}
\includegraphics[width=\linewidth]{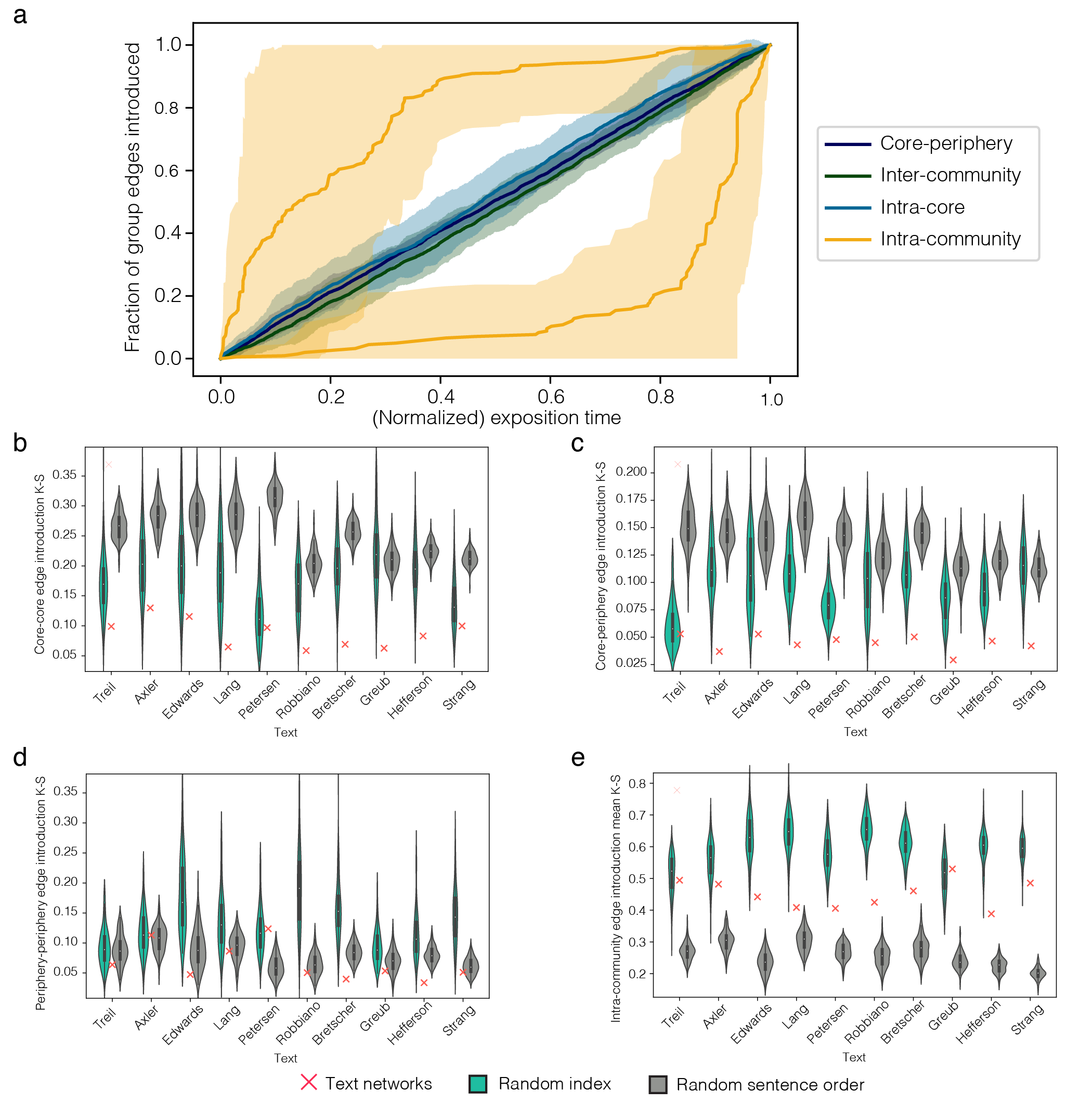}

\caption{Community development curves across texts, and associated K-S distance between community development curve types and the line $y=x$ across all texts and null ensembles. \emph{(a)} Mean $\pm$ 2 standard deviations of community development curves (fraction of edges within a particular group present at a particular normalized time in the exposition) across all texts, \emph{(b)} K-S distances for the core-core edge introduction curve, \emph{(c)} K-S distances for the core-periphery edge introduction curve, \emph{(d)} K-S distances for the periphery-periphery edge introduction curve, and \emph{(e)} mean K-S distances across intra-community edge introduction curves.}
\label{fig:supp_edgeintro}
\end{figure}

Similar to our analysis of the development of each text's core and periphery, we further wish to examine the development of the community structure in the semantic networks through the addition of edges between particular groups over the course of exposition. Specifically, we consider four edge types: `core-periphery' edges, or those connecting a core node with a periphery node; `intra-core' edges, connecting two core nodes; `inter-periphery' edges, connecting nodes in two different periphery communities; and `intra-community' edges, connecting two nodes in the same periphery community. We examine the relative introduction of each group of edge types by calculating, at each point in the texts' expositions, what fraction of edges in a particular group have been introduced. We show in Fig.~\ref{fig:supp_edgeintro} the mean $\pm$ 2 standard deviations of these group introduction curves across all texts; for the intra-community curves, we plot two examples: one of an early-introduced community, which attains a value near 1 indicating near-completion relatively quickly, and one of a late-introduced community, which takes longer to be fully developed, and remains closer to 0 throughout much of the text. 

Note that while the core-periphery, inter-community, and intra-core edge sets appear to be introduced steadily, showing little deviation from the diagonal $y=x$, which reflects constant introduction over time, the early and late intra-community examples shown have significant variability and deviate greatly from such a rule of constant introduction. We may quantify this behavior of deviation from constant introduction with the Kolmogorov-Smirnov (K-S) distance: in particular, for any of the edge group development curves $c(\cdot)$, we examine its K-S distance, or greatest vertical distance, to the line $y=x$ on the interval $(0, 1)$:
$$\text{K-S}(c) = \max_{t \in (0, 1)}|c(t)-t|.$$
Note that we chose our early- and late-introduced communities in Fig.~\ref{fig:supp_edgeintro} as those communities with the most positive and negative values of $c(t)-t$ on the interval $(0, 1)$, respectively. We plot the resulting K-S metrics for each edge group type across all texts and corresponding null models in Fig.~\ref{fig:supp_edgeintro}b-e. We find relative consistency across texts in relatively low K-S values for the intra-core, core-periphery, and inter-community groups, and notably, in many cases it appears as though the actual texts exhibit lower K-S values, and thus more constancy in edge introduction in these groups, than the bulk of the random index and random sentence order graphs (Fig.~\ref{fig:supp_edgeintro}b-d). Notably, we also observe that while many of the texts exhibit lower mean intra-community K-S values than the bulk of the random index networks, they also generally lie well above the distribution of values for the random sentence order null graphs. Thus, this pattern of findings suggests that while the texts generally exhibit significant variability in when intra-community edges are introduced during the exposition, the reordering of sentences that occurs in the random sentence order model disrupts this variability, causing a community's edges to, on average, be introduced in a more distributed fashion over the course of the reordered `exposition'. In turn, these findings suggest that the periphery communities extracted from the empirical networks do indeed represent distinct groups of related concepts that are localized in their position in text, as we might expect from a chapter focusing on a particular topic (Fig.~\ref{fig:supp_edgeintro}e).

\subsection*{Barcodes and Betti curves for all texts and null models}
\begin{figure}
 \centering
 \includegraphics[width=0.5\linewidth]{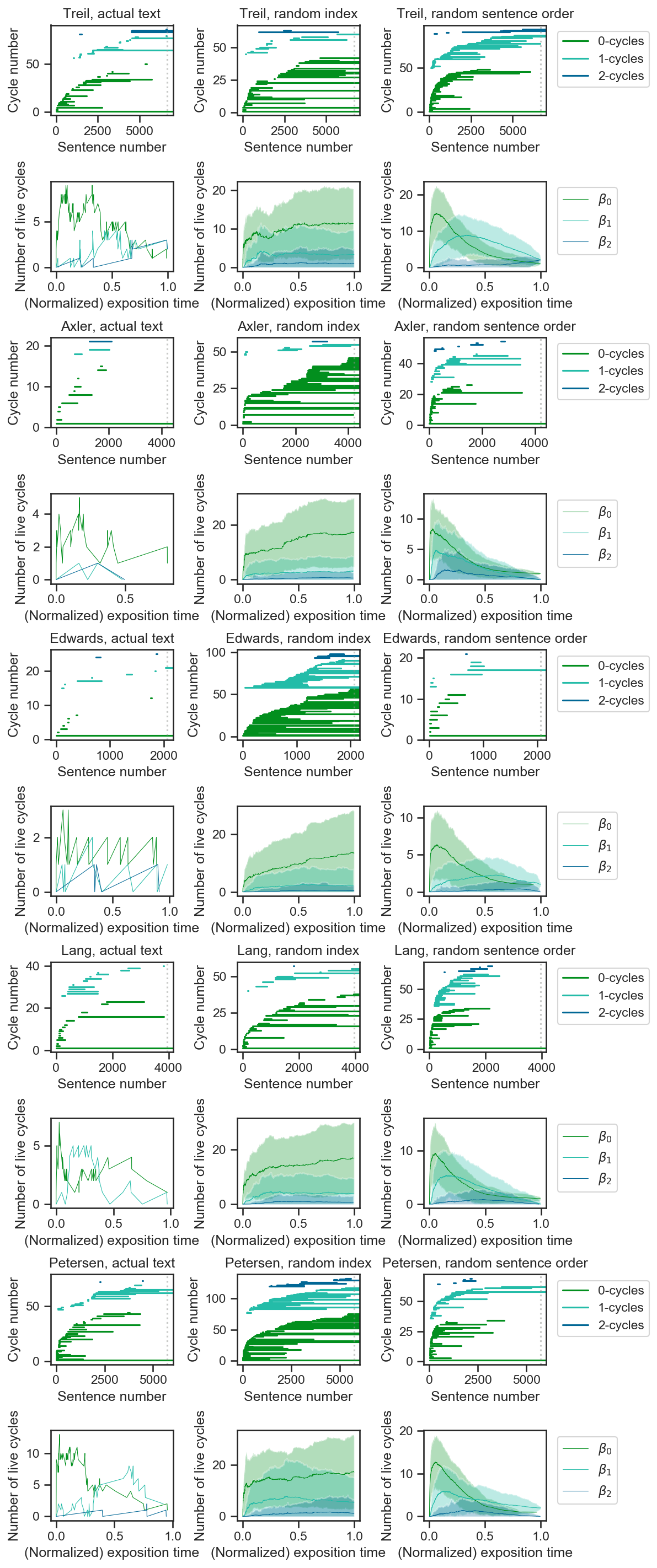}
 
\caption{Sentence-filtration barcodes and Betti curves for the first half of the texts. Each pair of rows shows an example barcode and Betti curves for a given text, with text results in the leftmost column and null models in the other columns.}
\label{fig:sbb_1}
\end{figure}

\begin{figure}
 \centering
 \includegraphics[width=0.5\linewidth]{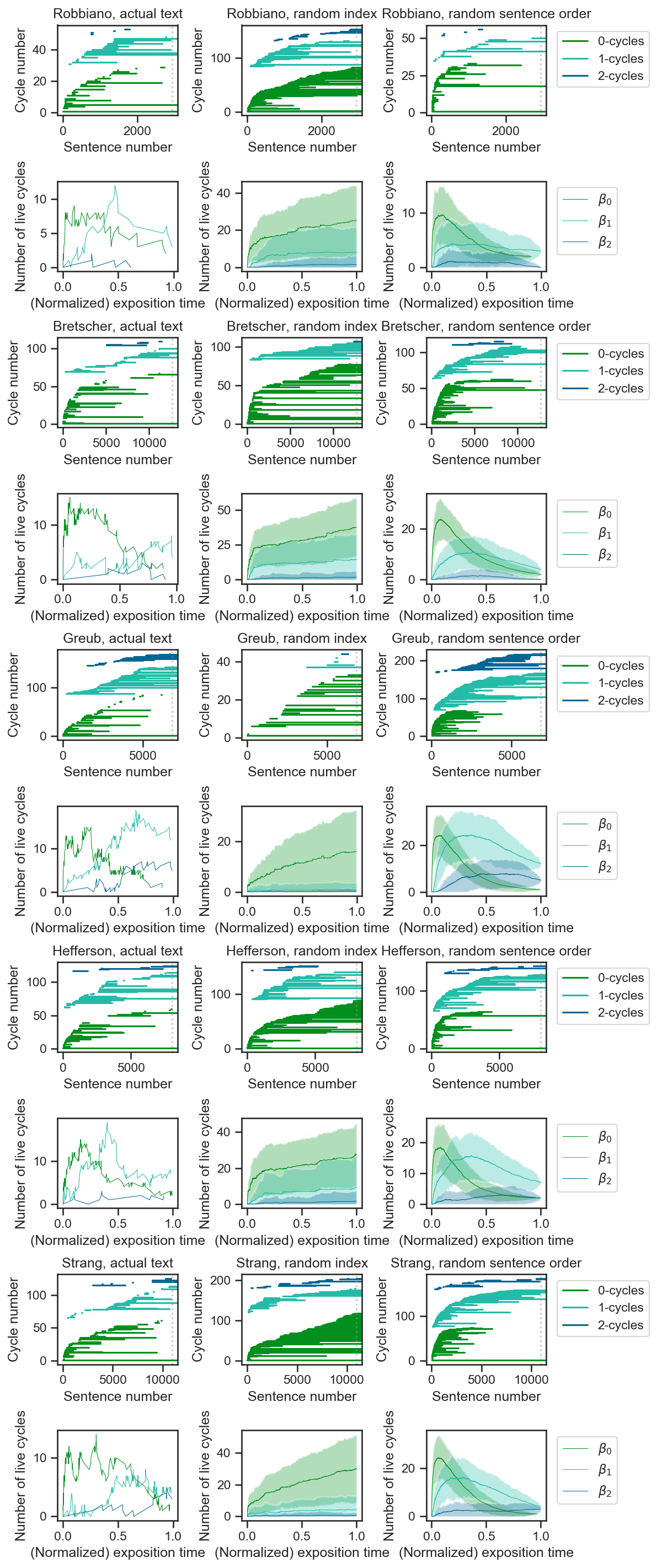}
 
\caption{Sentence-filtration barcodes and Betti curves for the second half of the texts. Each pair of rows shows an example barcode and Betti curves for a given text, with text results in the leftmost column and null models in the other columns.}
\label{fig:sbb_2}
\end{figure}

\begin{figure}
 \centering
 \includegraphics[width=0.95\linewidth]{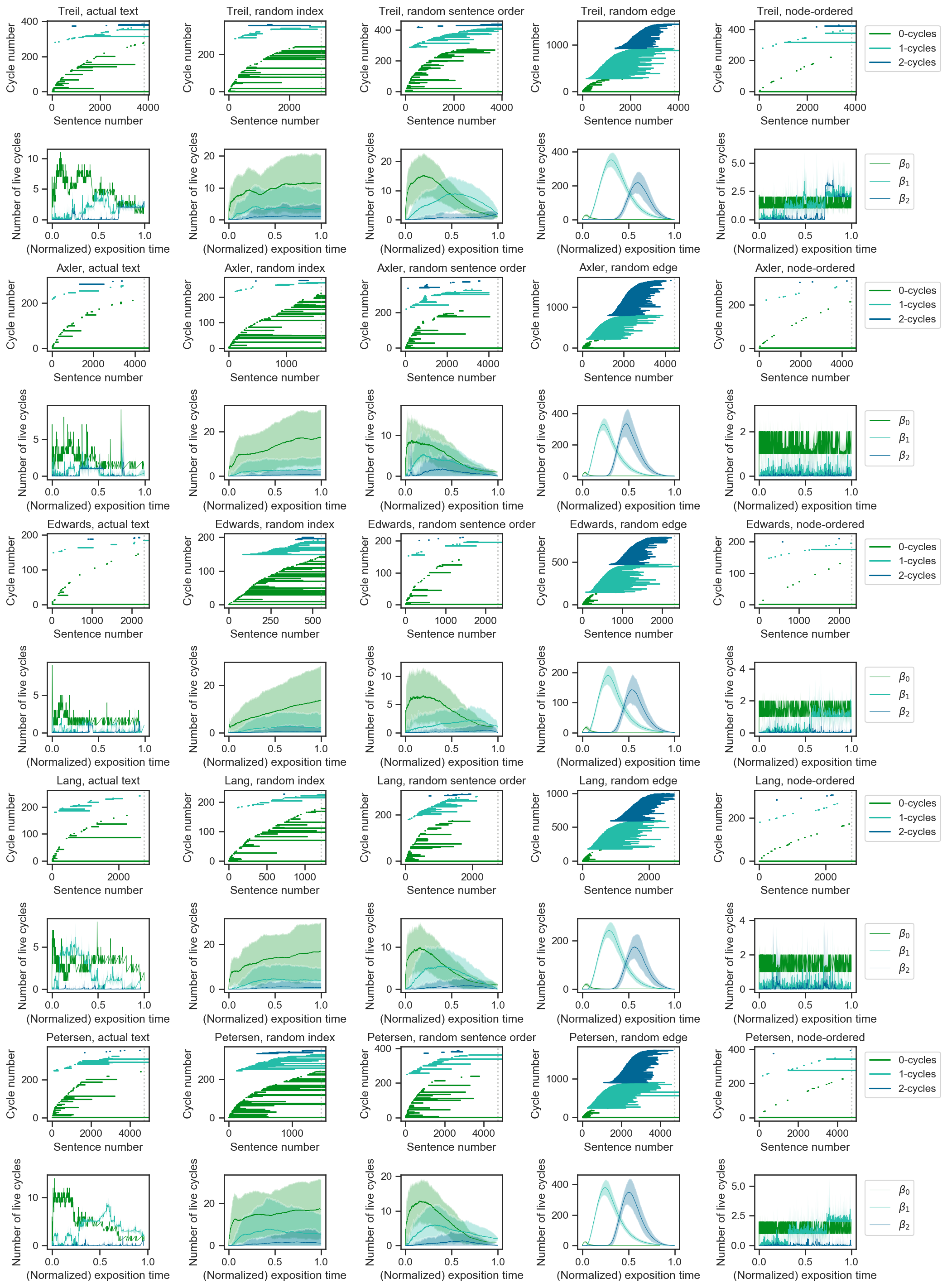}
 
\caption{OAAT barcodes and Betti curves for the first half of the texts. Each pair of rows shows an example barcode and Betti curves for a given text, with text results in the leftmost column and null models in the other columns.}
\label{fig:bb_1}
\end{figure}

\begin{figure}
 \centering
 \includegraphics[width=0.95\linewidth]{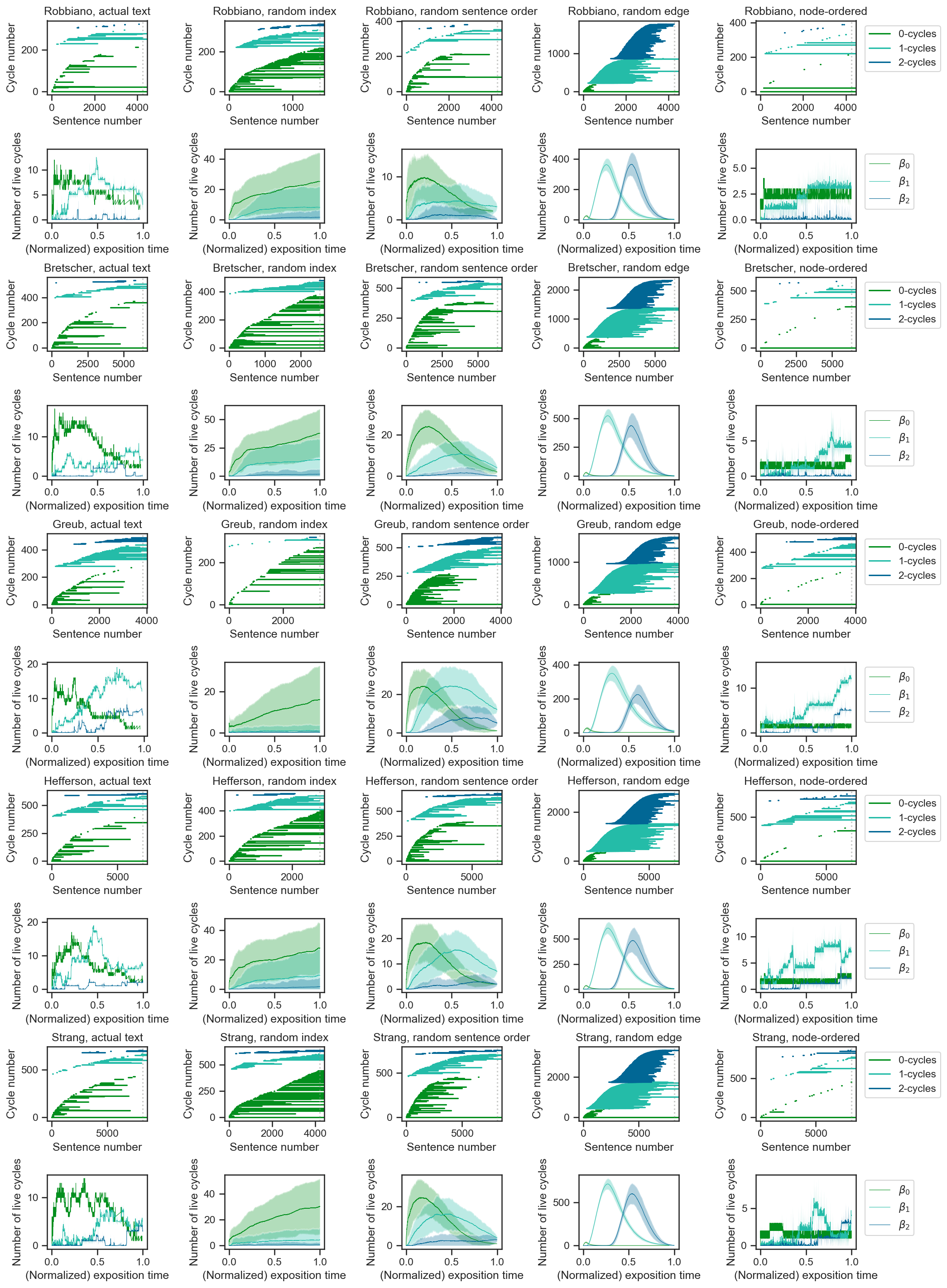}
 
\caption{OAAT barcodes and Betti curves for the second half of the texts. Each pair of rows shows an example barcode and Betti curves for a given text, with text results in the leftmost column and null models in the other columns.}
\label{fig:bb_2}
\end{figure}

For the barcodes and Betti curves of the sentence-granularity text filtration, random index model, and random sentence order model, see Figs.~\ref{fig:sbb_1}, \ref{fig:sbb_2}. For barcodes and Betti curves of the OAAT text filtration and all null ensembles, see Figs.~\ref{fig:bb_1}, \ref{fig:bb_2}.

\subsection*{Normalized average cycle lifetime for texts and all null ensembles}
\begin{figure}
\centering
\includegraphics[width=0.6\linewidth]{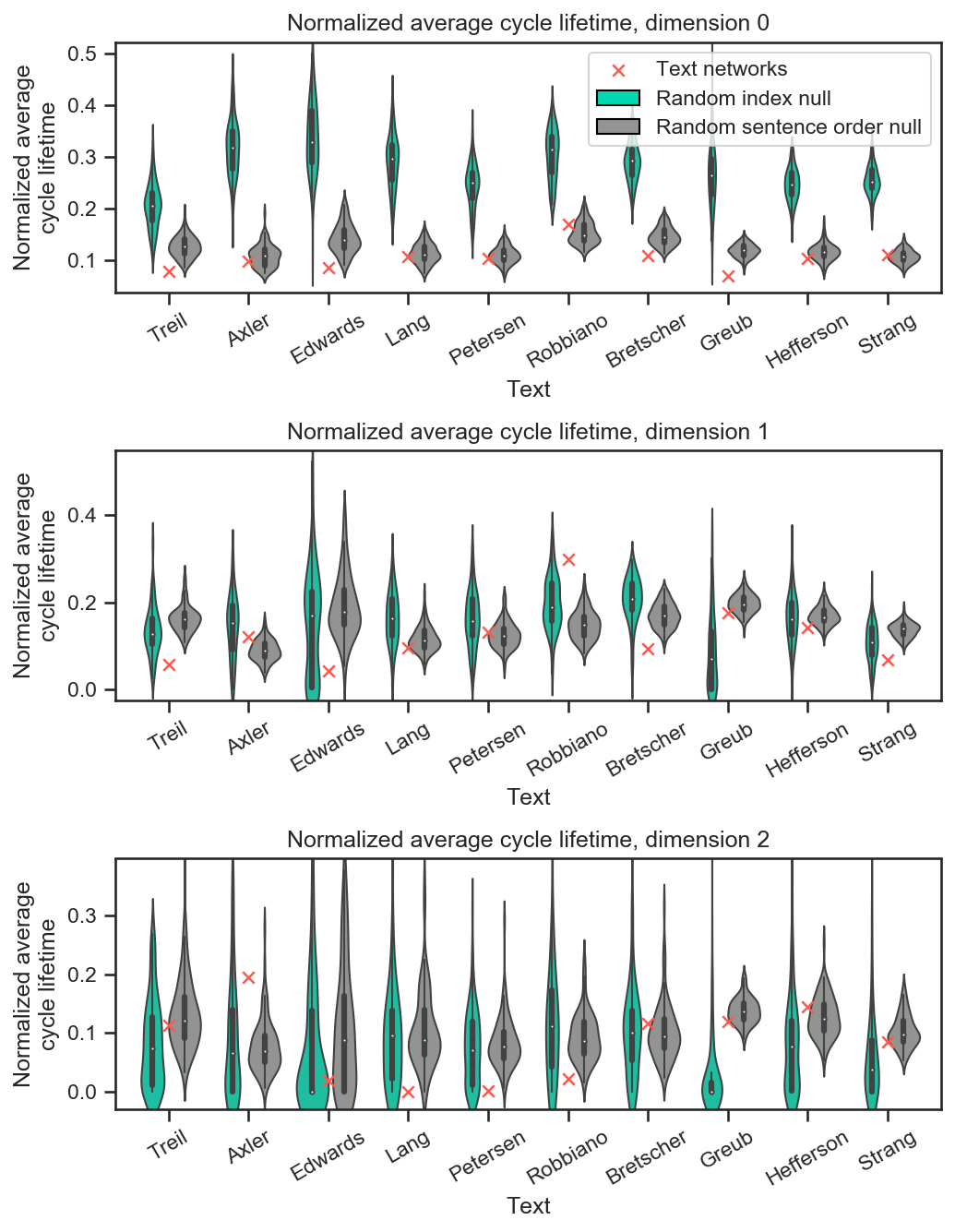}
\caption{Normalized average cycle lifetime for 0-, 1-, and 2-dimensional persistent homology across all texts' sentence-granularity filtrations and random index and random sentence order null models. From top to bottom: dimensions 0, 1, and 2.}
\label{fig:sentbarcodedens}
\end{figure}

\begin{figure}
\centering
\includegraphics[width=0.6\linewidth]{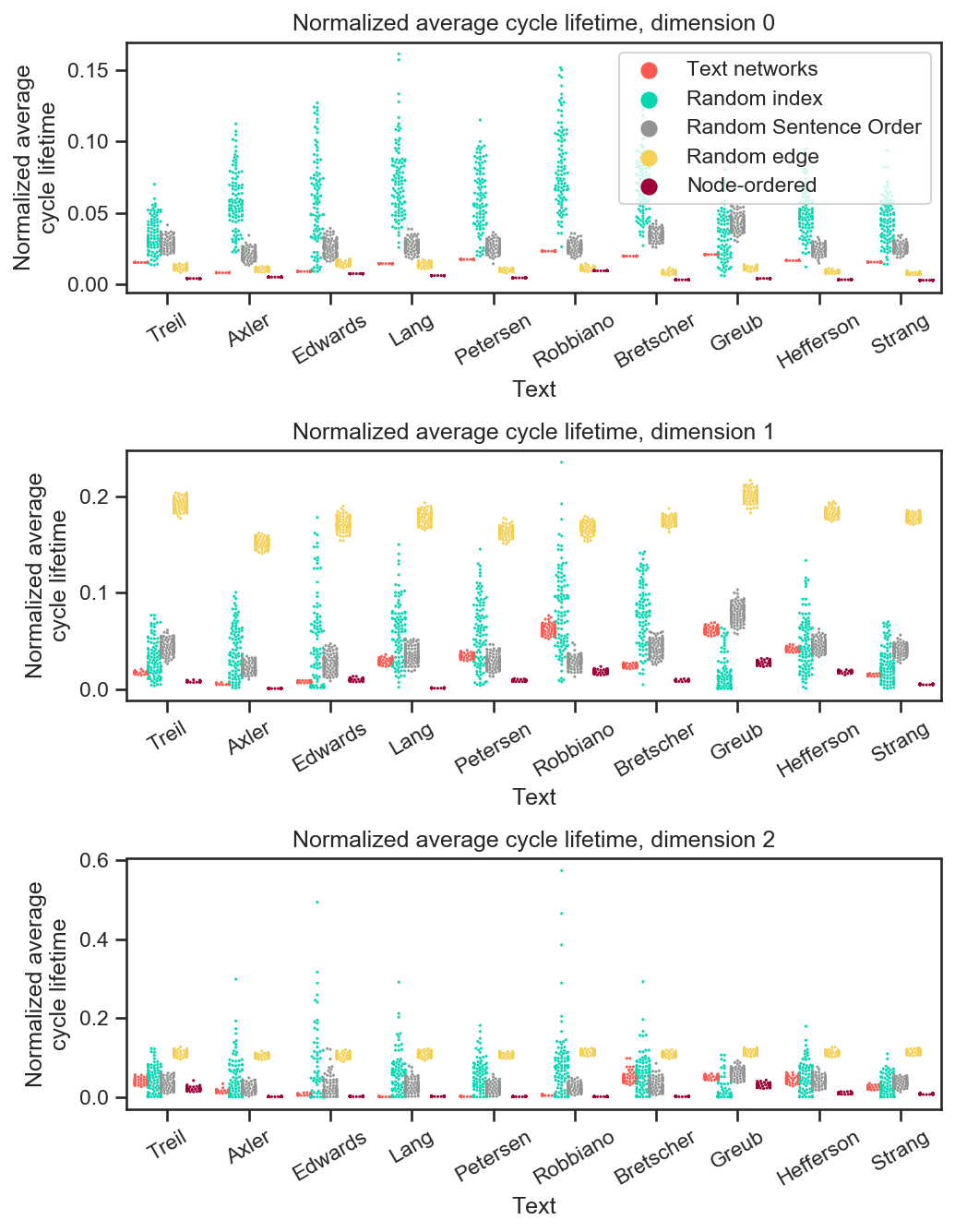}
\caption{Normalized average cycle lifetime for 0-, 1-, and 2-dimensional persistent homology across all texts' OAAT filtrations and all null models. From top to bottom: dimensions 0, 1, and 2.}
\label{fig:oaatbarcodedens}
\end{figure}

For normalized average cycle lifetimes of the sentence-granularity filtrations for the empirical texts, random index model, and random sentence order model, see Fig.~\ref{fig:sentbarcodedens}. For the normalized average lifetimes of the OAAT filtrations for the empirical texts and all null models, see Fig.~\ref{fig:oaatbarcodedens}.

\subsection*{Extended correlation analysis}

In the main text, we report results of a brief exploratory analysis assessing the relationship between structural features of exposition and community ratings of the textbooks from which the expositions are taken. Here, we provide the complete statistics for the Spearman and Pearson correlations between average rating on Goodreads and normalized average cycle lifetime (NACL) in Table~\ref{table:corr_pvalues}. We also note that while we consider average rating across editions, the default rating presented by Goodreads for textbooks, this metric should reasonably approximate the rating of each specific text edition we consider, since textbook editions tend to be similar.

\begin{figure}
 \centering
 \includegraphics[width=\linewidth]{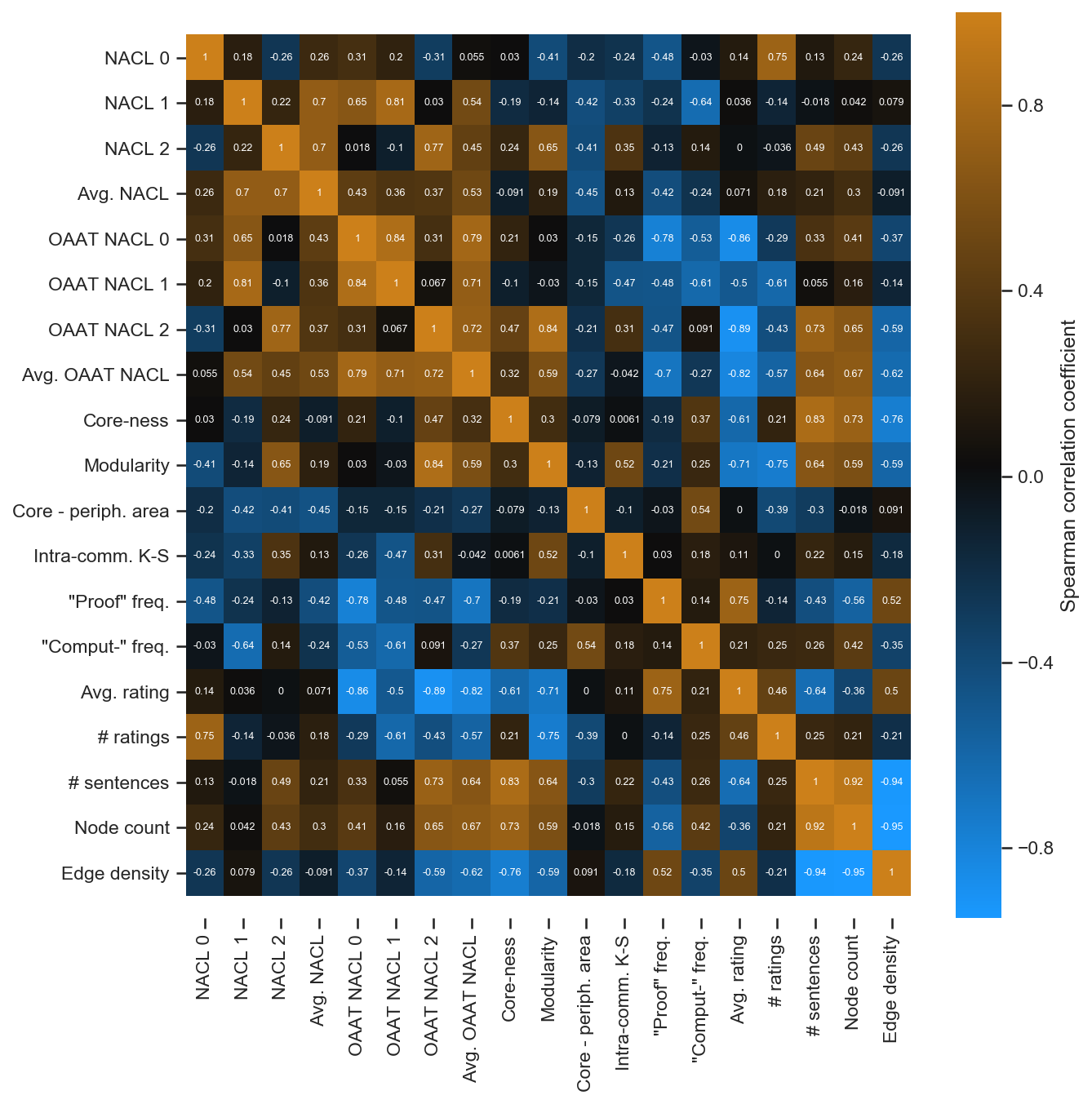}
 
\caption{Spearman correlation matrix for text features, including sentence- and OAAT-normalized average cycle lifetime (NACL), core-ness and modularity statistics, core - periphery area, intra-community edge development K-S, word frequencies, average text ratings and number of ratings, and text length, node count, and edge density. ``NACL $d$'' refers to NACL in dimension $d$.}
\label{fig:supp_corr}
\end{figure}

\begin{figure}
 \centering
 \includegraphics[width=\linewidth]{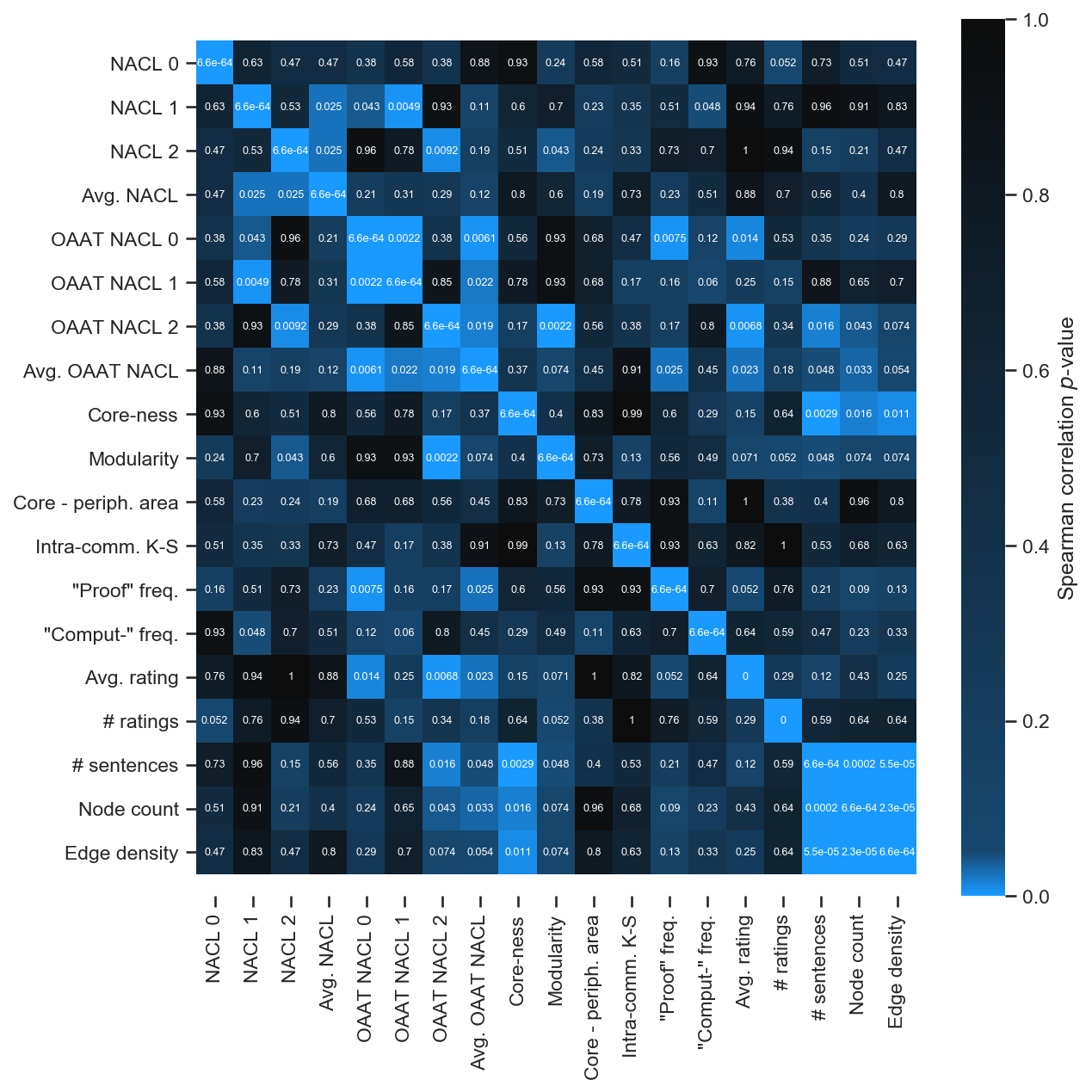}
 
\caption{Spearman correlation $p$-values for text features, including sentence- and OAAT-normalized average cycle lifetime (NACL), core-ness and modularity statistics, core - periphery area, intra-community edge development K-S, word frequencies, average text ratings and number of ratings, and text length, node count, and edge density. ``NACL $d$'' refers to NACL in dimension $d$.}
\label{fig:supp_p}
\end{figure}

We furthermore examine additional correlations between text features, both structural and otherwise, in Fig.~\ref{fig:supp_corr}, with associated $p$-values in Fig.~\ref{fig:supp_p}. Notably, while we observe correlations between average and dimension-2 OAAT NACL and both number of sentences and node count of each text, neither of the latter structural features are significantly correlated with the average text rating. Furthermore, though the number of ratings for each text is highly variable (Table~\ref{table:goodreads}), we find that this number does not significantly correlate with text rating (Spearman $\rho=0.464$, $p=0.294$). Finally, we find that both dimension-0 and average OAAT NACL are negatively correlated with the frequency of the word ``proof'' in the texts' sentences (Spearman $\rho=-0.782$, $p=0.00755$ and $\rho=-0.697$, $p=0.0251$, respectively), suggesting that theoretically-focused linear algebra texts might minimize the extent to which knowledge gaps are created and persist, compared to more applied texts. All correlations and $p$-values reported here and in the corresponding section of the main text were calculated using the Pingouin Python library, version 0.2.8 \cite{Vallat2018}.

\begin{table}
\centering
\begin{tabular}{|c|c|c|c|c|}
\hline
Correlate & Spearman corr. coef. & Spearman $p$-value & Pearson corr. coef. & Pearson $p$-value\\
\hline\hline
NACL, dim. 0 & 0.143 & 0.760 & 0.466 & 0.291\\\hline
NACL, dim. 1 & 0.036 & 0.939 & -0.334 & 0.464\\\hline
NACL, dim. 2 & 0.0 & 1.0 & -0.145 & 0.757\\\hline
Avg. NACL & 0.071 & 0.879 & -0.187 & 0.689\\\hline\hline
OAAT NACL, dim. 0 & -0.857 & 0.0137 & -0.821 & 0.0237\\\hline
OAAT NACL, dim. 1 & -0.500 & 0.253 & -0.575 & 0.177\\\hline
OAAT NACL, dim. 2 & -0.893 & 0.00681 & -0.846 & 0.0163\\\hline
Avg. OAAT NACL & -0.821 & 0.0234 & -0.828 & 0.0213\\\hline
\end{tabular}
\caption{Spearman and Pearson correlation coefficients and $p$-values for Goodreads ratings and normalized average cycle lifetimes (NACLs).}
\label{table:corr_pvalues}
\end{table}

\begin{table}
\centering
\begin{tabular}{|c|c|c|c|}
\hline
Text & Average Goodreads rating & Number of ratings \\
\hline\hline
Treil & 3.83 & 6\\\hline
Axler & 4.26 & 673\\\hline
Lang & 4.23 & 31\\\hline
Bretscher & 3.37 & 71\\\hline
Greub & 3.43 & 7\\\hline
Hefferson & 3.96 & 25\\\hline
Strang & 4.21 & 891\\\hline
\end{tabular}
\caption{Average rating and total number of ratings on Goodreads for texts with more than 5 ratings.}
\label{table:goodreads}
\end{table}

\section*{Supplementary Discussion}

\subsection*{The remarkable effectiveness of the random index null model}

Throughout our study, we have used the random index model as a null to examine how the results we obtain for the texts' actual semantic networks differ from what we might expect when simply calculating the co-occurrence networks and filtrations of a set of random words in a text. Notably, while most of our results have fallen at the extreme ends of the metric distributions exhibited by the random index ensemble, in some cases, such as in 1- and 2-dimensional normalized average cycle lifetime, the empirical filtrations demonstrate values that fall near the bulk of the corresponding random index ensemble results. The perspective that the null simply gives us a weighted network and filtration computed from the co-occurrence of a random set of words might be disheartening, as this could suggest that our results, rather than reflecting the meaningful structure of semantic networks of concepts elucidated by the textbooks, instead might simply reflect growing topologies that would be expected from any similar calculation of co-occurrence within a text. However, there is another lens through which we can view the random index null model; recalling that the random index sets are comprised of words not found within the stop word list, we might consider each random index graph as a semantic network itself. Certainly, the semantic features extracted through co-occurrence might not reflect the content which is the primary focus of the text, since the random index set might include non-mathematically-meaningful words. Even so, it is likely that some mathematical words will make their way onto the index set, and the remainder of the words are also meaningful in some way, since they are not stop words. Thus, the random index set actually may be viewed as an ensemble of semantic networks, each of which simply happens to have a different node set, and as such is extracting semantic information about the relationships between different words. In this frame of reference, the explanatory power of the random index model both makes sense, and should be expected. 

\end{document}